\documentclass[10pt,twocolumn,letterpaper]{article}

\usepackage[pagenumbers]{wacv} 

\usepackage{xcolor}
\definecolor{customblue}{RGB}{73, 123, 185}

\newcommand{\aquablue}[1]{{\color{blue}#1}}

\newcommand{\cblue}[1]{\textcolor{customblue}{#1}}

\newcommand{\benchmark}{\mbox{\textsc{VECTOR}}\xspace}
\newcommand{\method}{\mbox{\textsc{MECoT}}\xspace}
\newcommand{\nvideos}{\mbox{16.8k}\xspace}
\newcommand{\nquestions}{\mbox{4.8k}\xspace}

\usepackage{graphicx}
\usepackage{multirow}
\usepackage{rotating}
\usepackage{bbm}
\usepackage{pifont}

\usepackage{arydshln}

\definecolor{wacvblue}{rgb}{0.21,0.49,0.74}
\usepackage[pagebackref,breaklinks,colorlinks,allcolors=wacvblue]{hyperref}

\def\wacvPaperID{2224} 
\def\confName{WACV}
\def\confYear{2026}

\title{What Happens When: Learning Temporal Orders of Events in Videos}

\author{
Daechul Ahn$^{1,}\thanks{These authors contributed equally.}$\hspace{0.5em}
Yura Choi$^{2,*}$\hspace{0.2em}
Hyeonbeom Choi$^{1,*}$\hspace{0.2em}
Seongwon Cho$^{1}$\hspace{0.2em}
San Kim$^{1}$\hspace{0.2em}
Jonghyun Choi$^{1,}\thanks{JC is with ECE, ASRI and IPAI in SNU and a corresponding author.}$\vspace{0.3em} \vspace{0.5em}\\
\hspace{0.5em}$^1$Seoul National University\hspace{0.5em}$^2$Imperial College London\\
{\tt\small \{daechulahn,gusqja1228,seongwoncho,00sankim,jonghyunchoi\}@snu.ac.kr yc825@ic.ac.uk}
}

\begin{document}
\maketitle
\begin{abstract}
Video Large Multimodal Models (VLMMs) have shown impressive performance in video understanding, yet their ability to accurately capture the temporal order of multiple events remains underexplored. 
We interestingly observe that, even when video frames are scrambled, models perform very well on the existing benchmarks by comprehensive experiments.
This implies that VLMMs may not necessarily rely on accurate sequential processing of visual events, but instead depend on prior knowledge of typical scenarios to answer the question.
To benchmark temporal understanding capabilities in VLMMs, we propose \benchmark, designed to explicitly assess a model’s ability to identify the temporal order of events.
On this benchmark, we observe that various VLMMs often fail to understand the orders of events. 
To address this, we propose \method (Multi-Event instruction fine-tuning with Chain-of-Thought), which (1) trains models on detailed, event-by-event video descriptions and (2) using chain-of-thought prompts at inference to enhance temporal awareness. 
\method outperforms prior arts on \benchmark as well as improving performance on existing video benchmarks, implying effectiveness of temporal understanding.
We release our code, model and datasets\footnote{Project page:~\texttt{\href{https://dcahn12.github.io/BINDER}{\texttt{https://dcahn12.github.io/VECTOR}}}}.
\end{abstract}

\section{Introduction}
\label{sec:intro}

Humans naturally segment continuous visual sequences into discrete events and process them sequentially~\cite{Zacks2007EventPerception}.
This capability is crucial for cognition and event comprehension~\cite{Schank1977Scripts}.
Given these human capabilities, a question arises: `Do Video Large Multimodal Models (VLMMs) properly comprehend videos with multiple events in \emph{their temporal order}, similar to humans?'

\begin{figure}[t]
    \centering
    \includegraphics[width=1.0\linewidth]{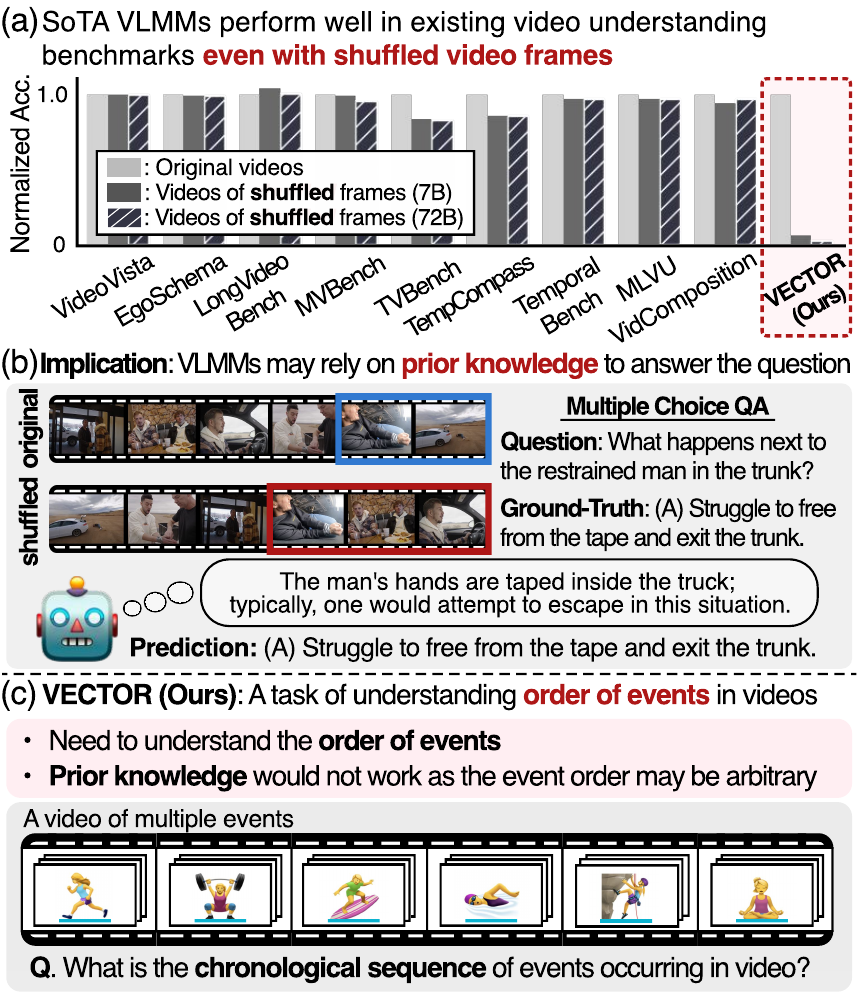}
    \vspace{-0.8em}
    \caption{\textbf{Limitations of existing video benchmarks and our \benchmark benchmark.}
    (a) Normalized accuracy (Acc.) indicates the ratio of a model's accuracy on frame-shuffled videos to its accuracy on original videos. 
    SoTA open-source VLMMs~\cite{li2024llavaov} (7B and 72B) achieve high accuracy on existing benchmarks but not on our proposed \benchmark.
    (b) We think existing benchmarks often allow models to succeed via prior knowledge of typical event scenarios, circumventing true temporal understanding presented in video. (c) In contrast, \benchmark explicitly evaluates event-order understanding independent of prior knowledge.
    }
    \label{fig:teaser}
    \vspace{-1.2em}
\end{figure}

Recent studies have explored this question by introducing various video benchmarks~\cite{li2024mvbench,mangalam2023egoschema,liu2024tempcompass,cores2024tvbench,wu2024longvideobench,li2024videovista,li2024vitatecs,tang2024vidcompostion} designed to assess the temporal understanding capabilities of VLMMs through video question answering tasks. 
However, our preliminary experiments reveal that state-of-the-art (SoTA) VLMMs~\cite{li2024llavaov} notably perform well even when video frames are randomly shuffled (Fig.\ref{fig:teaser}-(a)).
This indicates existing benchmarks often allow questions to be answered correctly without understanding temporal order in events.
We hypothesize that models leverage their \emph{prior knowledge} of typical event scenarios, inferring plausible contexts from isolated frames rather than actually analyzing temporal relationships explicitly depicted in videos~\cite{chen2024mecd,temporal_knowledge_1}.
Consequently, models may bypass temporal reasoning, resorting instead to common-sense shortcuts (see the supplementary material for detailed analysis).

\begin{figure}[t]
    \centering
    \includegraphics[width=1.0\linewidth]{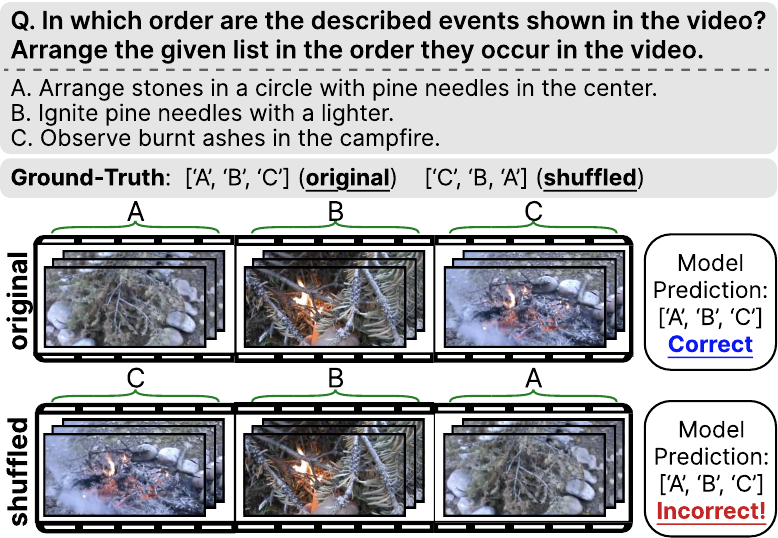}
    \caption{\textbf{Biased prediction on event-ordering task}.  The model correctly predicts the event order [`A', `B', `C'] for the original video (top). However, when events A and C are swapped (bottom), it maintains the same prediction, revealing its reliance on \emph{plausible scenario} over visual temporal information.
    }
    \label{fig:chronological_natural}
\end{figure}

While such prior  can facilitate video understanding~\cite{chen2024mecd}, excessive reliance on it may lead VLMMs to produce biased interpretations of video.
We empirically demonstrate this bias in event-ordering tasks with instructional videos~\cite{Zala2023HiREST,chen2024mecd} (detailed quantitative analysis in~$\S$\ref{subsec:diagnose_shortcut}).
For example, in Fig.\ref{fig:chronological_natural}, GPT-4o~\cite{gpt-4o} correctly predicts the sequence of a campfire-making video but retains the same prediction even when events \texttt{A} and \texttt{C} are swapped.
Namely, it follows the \emph{plausible progression} rather than recognizing the altered order shown in the video.
This highlights the need for VLMMs to develop robust temporal reasoning capabilities that function \emph{beyond} common-sense priors for comprehensive spatiotemporal understanding.

To address this limitation of temporal understanding, we introduce \textbf{\benchmark} (\textbf{V}isual \textbf{E}vent \textbf{C}hronology and \textbf{T}emporal \textbf{O}rder \textbf{R}easoning), a diagnostic benchmark complementary to existing video benchmarks.
\benchmark explicitly evaluates event-order comprehension without reliance on common-sense priors. 
To achieve this, we construct synthetic videos by concatenating distinct action clips from Kinetics~\cite{smaira2020short_kinetics}, deliberately inserting abrupt transitions to prevent reliance on prior knowledge or plausible scenarios (Fig.~\ref{fig:teaser}-(c)). 
Unlike real-world videos with predictable event sequences, our controlled videos intentionally break common-sense expectations, requiring models to explicitly analyze temporal relationships.
Clear event segmentation further enhances interpretative clarity, resembling rapid context shifts common in edited media like news highlights or sports recaps.

Our comprehensive evaluation of various open-source and proprietary VLMMs on \benchmark shows that contemporary models often struggle to grasp temporal order across multiple video events. 
To effectively reason about temporal structures, models must learn how individual events relate to each other and understand the temporal progression underlying event sequences~\cite{zhou2018temporal,guo2025trace,cheng2025tempura}. 
Motivated by this, we propose \textbf{\method} (\textbf{M}ulti-\textbf{E}vent instruction fine-tuning with \textbf{C}hain-\textbf{o}f-\textbf{T}hought), which combines detailed event-by-event instruction fine-tuning with structured chain-of-thought reasoning during inference to explicitly reason about event progression. 
\method achieves notable improvements on \benchmark and demonstrates enhanced performance on various existing video benchmarks.
\noindent Our contributions are as follows: 
\begin{itemize}[noitemsep,topsep=0pt,partopsep=0pt]
\item A comprehensive analysis of a wide range of VLMMs and empirically reveal their deficiency in temporal understanding in videos.
\item Proposing \benchmark, a diagnostic benchmark to comprehensively evaluate VLMMs' event-wise and pattern-wise temporal order understanding on multi-event videos.
\item Proposing a simple yet effective method (\method) that combines fine-tuning on event-by-event video descriptions with a chain-of-thought reasoning mechanism, improving temporal understanding on both \benchmark and other video benchmarks.
\end{itemize}

\section{Related Work}
\label{sec:related_work}
\paragraph{Video large multimodal models (VLMMs).}
Recent research integrates visual encoders with large language models (LLMs) to enhance multimodal understanding~\cite{Maaz2023VideoChatGPT,shen2024longvu,chen2024internvl,Qwen2VL,aria,zhang2024llavanextvideo}.  
VLLMs typically use image-based encoders like CLIP~\cite{radford2021learning} for frame-level processing, aligning visual and language embeddings via projection layers~\cite{li2024llavaov,Maaz2023VideoChatGPT,Maaz2024VideoGPT+,damonlpsg2024videollama2}.  
Video-ChatGPT~\cite{Maaz2023VideoChatGPT} employs spatial and temporal pooling, while VideoGPT+~\cite{Maaz2024VideoGPT+} combines image and video encoders.  
BT-Adapter~\cite{liu2024bt} enhances temporal modeling with branching adapters.  
LLaVA-OneVision~\cite{li2024llavaov} uses curriculum learning for image-to-video understanding, and LLaMA-VID~\cite{li2025llama} optimizes video processing with context tokens per frame.  
For longer videos, PLLaVA~\cite{xu2024pllava} preserves image understanding via spatial-temporal pooling,  
PPLLaVA~\cite{liu2024ppllava} introduces instruction-aware pooling for token compression,  
and LongLLaVA~\cite{wang2024longllavascalingmultimodalllms} integrates Mamba~\cite{gu2023mamba} for efficient long-context processing.

\paragraph{Temporal understanding benchmarks for VLMMs.}
While VLMMs have demonstrated strong performance on various video understanding datasets~\cite{li2024llavaov,damonlpsg2024videollama2}, reliably assessing their \emph{temporal} understanding remains challenging. Existing benchmarks employ three primary evaluation strategies: requiring minimal viewing length to reduce single-frame shortcuts~\cite{mangalam2023egoschema,wu2024longvideobench}, focusing on specific temporal skills such as speed, direction, and event linkage~\cite{liu2024tempcompass,cores2024tvbench,chen2024rextime,zhang2024vinoground,velociti}, and evaluating sequential change detection across clips~\cite{cai2024temporalbench,mlvu,li2024videovista}. However, as illustrated in Fig.~\ref{fig:teaser}-(a) and (b), VLMMs often succeed even with shuffled frames, underscoring the need for more diagnostic evaluation methods.
Our benchmark distinguishes itself through a novel approach: we concatenate multiple unrelated single-event clips into synthetic multi-event videos, compelling models to interpret distinct event sequences without relying on prior knowledge or common progressions. 
Further, we replace previous multiple-choice questions with formatted sequence generation tasks that require explicit enumeration and ordering of events. This design prevents shortcut-based guessing and enables deeper diagnostic analysis. 

Among existing approaches, TempCompass~\cite{liu2024tempcompass} and MLVU~\cite{mlvu} are the most similar to \benchmark. 
TempCompass evaluates temporal perception but is limited to short videos with a single event, which often allows many questions to be answered from only a few frames.
MLVU~\cite{mlvu} assesses action order in long videos by inserting target actions into lengthy backgrounds, often emphasizing context searching over systematic temporal sequence examination. 
In contrast, \benchmark constructs sequences composed entirely of distinct events, requiring direct reasoning over multiple successive events. 
Moreover, \benchmark evaluates both full-sequence and sub-sequence ordering with detailed metrics beyond exact match, providing a more comprehensive assessment of temporal understanding.
See the supplementary material for detailed comparisons.

\section{\benchmark: Comprehending Temporal Order of Events in Videos}
\label{sec:vectorbench}

\subsection{Diagnosing the Prior-Knowledge Bias}
\label{subsec:diagnose_shortcut}
Human video understanding combines \emph{visual perception} with \emph{prior knowledge}~\cite{lake2017building}. 
For example, seeing someone enter a kitchen and reach for ingredients evokes expectations of cooking based on common event schemas~\cite{Zacks2007EventPerception}. 
However, in VLMMs, such priors may serve as \emph{shortcuts}, hindering genuine comprehension of temporal sequences~\cite{battaglia2018relational}.

We examine VLMMs' reliance on prior knowledge in event-ordering tasks through controlled experiments on three proprietary models (Claude-3.5~\cite{claude}, GPT-4o~\cite{gpt-4o}, Gemini~\cite{gemini}), using two procedural and instructional datasets with segmented events (3--6 per video), each accompanied by scripted captions~\cite{chen2024mecd,Zala2023HiREST}. 
Models chronologically order events under original (\textbf{original}) and event-wise shuffled (\textbf{shuffled}) conditions as in Fig.~\ref{fig:chronological_natural} (details in the supplementary material).

\paragraph{Biased ratio.}
To quantitatively analyze this shortcuts, we define \emph{biased ratio} ($\eta$) as follows.

\begin{equation}
\eta = \frac{|\{x : (P_{\text{o}}(x) = P_{\text{s}}(x)) \wedge C_{\text{o}}(x) \wedge \neg C_{\text{s}}(x)\}|}{|\{x : C_{\text{o}}(x) \wedge \neg C_{\text{s}}(x)\}|,}
\end{equation}
where $x$ is a multi-event video, $P_{o}(x)$ and $P_{s}(x)$ denote VLMM's predicted event orderings for original and shuffled videos respectively, while $C_{o}(x)$ and $C_{s}(x)$ indicate prediction correctness for original and shuffled videos.
In words, among videos for which the model’s prediction is correct when events are in the original order but incorrect when events are shuffled, $\eta$ measures the fraction that receive the \emph{same} predicted order in both conditions.

This metric reveals how often models favor causally plausible orderings over visual temporal evidence, indicating their reliance on prior knowledge rather than observed temporal information.
Empirically, \(\eta\) exceeds 78\% for all three proprietary models in Tab.~\ref{tab:caption_re_order_natural} across both datasets, indicating a reliance on prior knowledge rather than on actual temporal evidence, as illustrated in Fig.~\ref{fig:chronological_natural}.

\subsection{\benchmark Benchmark}
\label{subsec:vector_overview}

\begin{figure*}[t]
    \centering
    \includegraphics[width=1\linewidth]{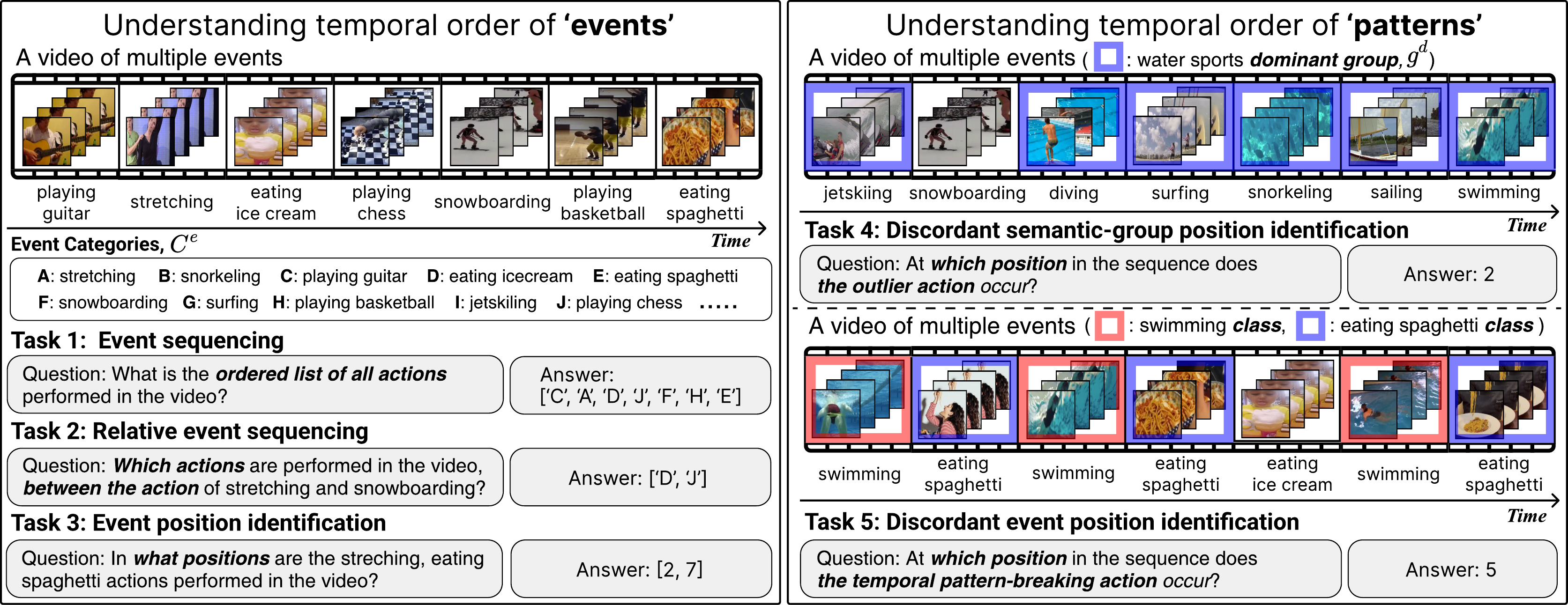}
    \vspace{-1.5em}
    \caption{\textbf{Overview of the \benchmark benchmark}. Two evaluation groups: \textbf{Left}: Event‐level tasks assess event sequencing, relative sequencing, and position identification. \textbf{Right}: Pattern‐level tasks group events into semantic categories or recuring patterns, requiring models to detect anomalous event's position within sequences.}
    \label{fig:overview}
    \vspace{-1.0em}
\end{figure*}

To investigate VLMMs' temporal understanding independent of prior knowledge, we introduce \benchmark, which includes two task groups: (1) \textbf{Event ordering}, focusing on enumerating \emph{individual} events, and (2) \textbf{Pattern ordering}, identifying \emph{semantic patterns} across events (\eg, repeated sequences or thematic groups) as shown in Fig.~\ref{fig:overview}. 
Each task has two difficulty levels: fewer events (\textbf{L1}) and more events (\textbf{L2}). 
We create synthetic multi-event videos by temporally concatenating unrelated short clips from existing datasets, preventing reliance on prior knowledge. 
The dataset comprises \nquestions QA pairs with \nvideos action videos across five sub-tasks; detailed prompts and construction procedure are provided in the supplementary material.

\begin{table}[t]
    \centering   
    \setlength{\tabcolsep}{3.2pt}
    \resizebox{1\linewidth}{!}{
    \begin{tabular}{lcccccc}
        \toprule
        \multirow{2.5}{*}{Models} & \multicolumn{3}{c}{MECD~\cite{chen2024mecd}} & \multicolumn{3}{c}{HiREST~\cite{Zala2023HiREST}} \\
        \cmidrule(lr){2-4} \cmidrule(lr){5-7}
        & Org. & Shuf. & $\eta$ (\%) & Org. & Shuf. & $\eta$ (\%)\\
        \cmidrule(r){1-1} \cmidrule(lr){2-4} \cmidrule(lr){5-7}
        Claude-3.5-Sonnet & 37.63 & 1.77 & 85.71 & 26.69 & 2.70 & 78.21 \\
        Gemini 1.5 Pro & 32.60 & 2.58 & 93.22 & 25.34 & 2.70 & 90.67 \\
        GPT-4o & 40.82 & 2.06 & 80.00 & 35.82 & 1.69  & 83.50 \\
        \bottomrule
    \end{tabular}
    }
    \vspace{-0.65em}
    \caption{
    \textbf{Event sequencing task on videos with original and shuffled event order.}
    Exact match accuracy of event sequencing on the original video (`Org'), against a randomized order of events (`Shuf'). We report the `\textbf{biased ratio}'($\eta$) to quantify the model's reliance on prior knowledge rather than actual temporal cues.
    }
    \vspace{-1.0em}
    \label{tab:caption_re_order_natural} 
\end{table}

\vspace{-0.5em}
\paragraph{Understanding temporal order of `events'.}
We define an \emph{event} as a set of contiguous frames depicting a single human action, without overlap between consecutive events.
A multi-event video $V$ comprises an ordered set of events:
$V = (e_k)_{k=1}^{N_e}$, where $e_k$ is the $k$-th event and $N_e$ is the total number of events.
Event-level tasks assess a model’s ability to correctly enumerate or locate these events in temporal order.
Each event belongs to one of $N_c=20$ predefined categories $\mathcal{C}^e = \{{c_i}\}_{i=1}^{N_c}$, chosen based on VLMMs' sufficient single-action recognition performance, as shown in Fig.~\ref{fig:overview} (left); details in the supplementary material.

\vspace{-0.5em}
\paragraph{Task 1: Event sequencing.}
The first task evaluates the model’s ability to enumerate all events chronologically from the candidate set $\mathcal{C}^e$. 
We define two difficulty levels: \textbf{L1} (four events, $N_e=4$) and \textbf{L2} (eight events, $N_e=8$). 
L1 contains shorter sequences, while L2 demands tracking longer event histories, increasing complexity.

\paragraph{Task 2: Relative event sequencing.}
To assess more targeted temporal understanding ability, this task focuses on a specific sub-range of the video, \ie, first, VLMMs localize the two query events, then list all events occurring between them. 
Given two query events $(e_{q_i})$ and $(e_{q_j})$ with $q_i < q_j$, the model must identify all events $\{e_k\}$ such that $q_i < k < q_j$. 
As in Task~1, we use $\mathcal{C}^e$ as candidates and offer two task difficulty levels (L1, L2).

\paragraph{Task 3: Event position identification.}
This task evaluates the ability of VLMMs to determine the relative temporal positions of queried events within a sequence by identifying their indices, requiring precise location of \emph{when} specific events occur.
The model is given a query set $\{e_{q_i}\}_{i=1}^{N_q}$ and must output the positions $\{q_i\}_{i=1}^{N_q}$ of these events in the sequence. 
We vary $N_q$ from 1 to 3 to adjust difficulty, and offer two task difficulty levels as in Task~1.

\paragraph{Understanding temporal order of `patterns'.}
We define a \emph{pattern} as a high-level temporal or conceptual structure (semantic or repetitive) shared by most events in a video. 
Models must identify events that deviate from these patterns, requiring deeper temporal comprehension \emph{beyond} individual event recognition. 
We group event categories ($\mathcal{C}^e$) into 50 semantic groups $\mathcal{G}$, each containing about seven related actions (\eg, `water sports' includes `diving' and `swimming'; details in the supplementary material).


\vspace{-0.5em}
\paragraph{Task 4: Discordant semantic-group position identification.}
In this task, we seek to identify the single \emph{semantic outlier}---an event that does not belong to the dominant semantic group in the sequence, requiring high-level grouping to detect where the pattern breaks.
Formally, we define a video’s event sequence as \(\{e_1, e_2, \dots, e_m\}\), each event \(e_i\) belonging to a group \(g_{e_i} \in \mathcal{G}\). 
The \emph{dominant group} \(g^d\) is:
\begin{equation}
    g^d = \underset{g \in \mathcal{G}}{\arg\max}
    \sum_{i=1}^{m} \mathbbm{1}[g_{e_i} = g].
\end{equation}
There is exactly \emph{one outlier} event \(e_k\) such that \(g_{e_k} \neq g^d\) and \(g_{e_i} = g^d\) for all \(i \neq k\). 
The task is to locate this outlier position $k$. 
We offer two difficulty levels: \textbf{L1} with four events $(N_e=4)$ and \textbf{L2} with eight events $(N_e=8)$. 

\paragraph{Task 5: Discordant event position identification}
This task requires recognizing temporal patterns and identifying a single event $x$ disrupting the repeating sequence $(s_i)_{i=1}^m$, where $s_i$ is the $i^\text{th}$ event in the pattern. 
For example, in $V = (s_1, s_2, s_1, x, s_2, s_1, s_2)$, the pattern $(s_1, s_2)$ is repeated, but \(x\) at position 4 breaks this repetition. The model must locate the outlier $x$. 
We control difficulty by varying pattern length ($m \in \{2,3\}$) and repetitions (two for \textbf{L1}, three for \textbf{L2}). 
As illustrated in Fig.~\ref{fig:overview} (right), a sequence alternating between `swimming' and `eating spaghetti' ($m=2$) becomes disrupted by `eating ice cream,' requiring models to detect this discordant event.

\begin{figure}[t]
    \centering
    \includegraphics[width=\linewidth]{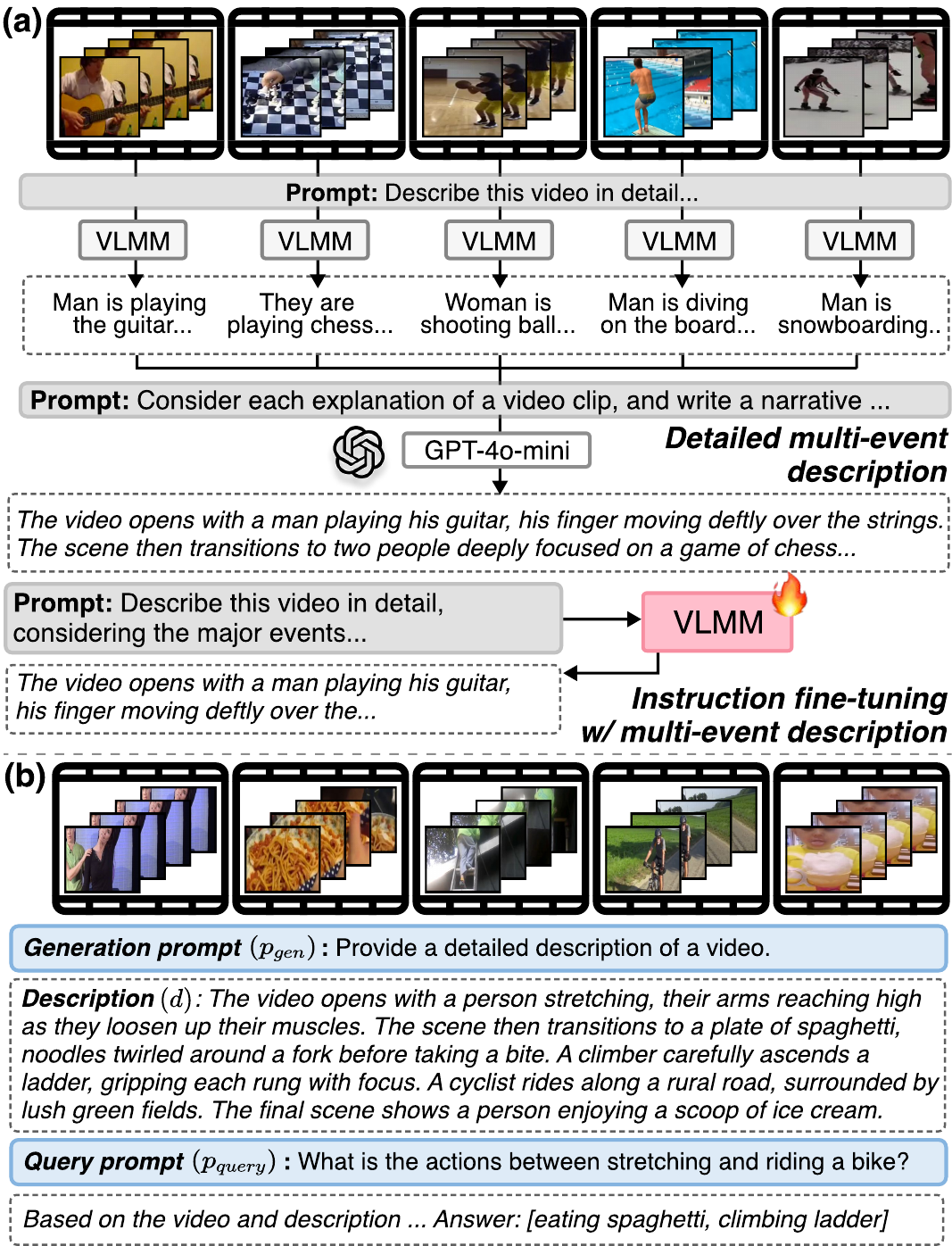}
    \vspace{-1.0em}
    \caption{\textbf{Overview of \method}. \method includes two stages: (a) Instruction fine-tuning on detailed multi-event descriptions to enhance event-level understanding. (b) Chain-of-thought (CoT) inference, where models generate structured narratives for videos, enabling explicit temporal reasoning before answering questions.}
    \label{fig:temp_inst_tune}
    \vspace{-1.0em}
\end{figure}

\begin{table*}[t]
    \centering
    \resizebox{\linewidth}{!}{
    \begin{tabular}{lcccccccccccccc}
        \toprule
        &&& 
        & \multicolumn{8}{c}{Open-source models}  
        & \multirow{2.75}{*}{\shortstack{\textbf{\method}\\({Ours})}}
        & \multicolumn{2}{c}{Proprietary models} \\
        \cmidrule(lr){5-12} \cmidrule(lr){14-15}
        Exact Match (EM)~$\uparrow$ & Lvl & Chance & Human
            & LongVA & Oryx & VILA-1.5 & V-LM2 & LongLV & LongVU & NVILA & LV-OV
            & 
            & GPT-4o & Gem.1.5P \\
        \midrule
        \addlinespace[5pt]
        \multicolumn{11}{l}{\textsc{\textbf{Task 1}}: \textsc{Event sequencing}} \\
        \midrule
        \multirow{2}{*}{\raisebox{-1.0ex}{Full-sequence ordering}}
        & \textsc{L1} & 0.00 & 85.00 
            & 1.33 & 0.00 & 0.33 & 2.33 & 0.00 
            & 12.67 & 20.00 &\underline{23.00} & \textbf{41.67} 
            & 80.00 & 78.00 \\[0.5ex]
        \cdashline{2-15}
        \\[-2ex]
        & \textsc{L2} & 0.00 & 80.00 
            & 0.00 & 0.00 & 0.00 & 0.00 & 0.00 
            & 0.00 & \underline{0.33} & \underline{0.33} & \textbf{4.33}
            & 51.67 & 59.33 \\
        \midrule
        \multicolumn{11}{l}{\textsc{\textbf{Task 2}}: \textsc{Relative event sequencing}} \\
        \midrule
        \multirow{2}{*}{\raisebox{-1.0ex}{Sub-sequence ordering}}
        & \textsc{L1} & 2.94 & 90.00 
            & 11.33 & 0.67 & 4.67 & 24.67 & 9.33 
            & 24.33 & 29.00 & \underline{40.33} & \textbf{55.67}
            & 83.33 & 80.67 \\[0.5ex]
        \cdashline{2-15}
        \\[-2ex]
        & \textsc{L2} & 0.98 & 85.00 
            & 4.33 & 0.33 & 1.67 & 3.00 & 2.67 
            & 5.33 & 7.00 & \underline{8.00} & \textbf{8.33}
            & 53.67 & 56.33 \\
        \midrule
        \multicolumn{11}{l}{\textsc{\textbf{Task 3}}: \textsc{Event position identification}} \\
        \midrule
        \multirow{2}{*}{\raisebox{-1.0ex}{Single event detection}}
        & \textsc{L1} & 25.00 & 100.00
            & 24.33 & 32.00 & 28.00 & 22.00 & 19.00 
            & 53.00 & \underline{60.33} & 37.00 & \textbf{93.33}
            & 75.67 & 82.67 \\[0.5ex]
        \cdashline{2-15}
        \\[-2ex]
        & \textsc{L2} & 12.50 & 85.00 
            & 15.67 & 12.00 & 12.50 & 12.67 & 6.67 
            & 27.00 & \underline{31.33} & 20.67 & \textbf{71.33}
            & 46.00 & 51.00 \\
        \midrule
        \multirow{2}{*}{\raisebox{-1.0ex}{Double event detection}}
        & \textsc{L1} & 8.33 & 95.00 
            & 8.00 & 8.67 & 11.33 & 9.00 & 6.33 
            & \underline{17.33} & 13.00 & 17.00 & \textbf{48.67}
            & 47.33 & 64.33 \\[0.5ex]
        \cdashline{2-15}
        \\[-2ex]
        & \textsc{L2} & 1.79 & 85.00 
            & 1.00 & 1.67 & 3.57 & 1.33 & 1.67 
            & 2.67 & 2.00 & \underline{6.00} & \textbf{43.67}
            & 14.00 & 30.33 \\
        \midrule
        \multirow{2}{*}{\raisebox{-1.0ex}{Triple event detection}}
        & \textsc{L1} & 4.17 & 85.00 
            & 9.33 & 8.33 & 5.33 & 7.00 & 0.67 
            & 14.67 & 17.67 & \underline{31.67} & \textbf{48.00}
            & 41.00 & 45.33 \\[0.5ex]
        \cdashline{2-15}
        \\[-2ex]
        & \textsc{L2} & 0.30 & 80.00 
            & 1.00 & 1.00 & 1.78 & 0.33 & 0.00 
            & 0.67 & 0.00 & \underline{2.33} & \textbf{20.00}
            & 7.33 & 26.67 \\
        \midrule
        \multicolumn{11}{l}{\textsc{\textbf{Task 4}}: \textsc{Discordant semantic-group position identification}} \\
        \midrule
        \multirow{2}{*}{\raisebox{-1.0ex}{Single anomaly}}
        & \textsc{L1} & 25.00 & 100.00
            & 26.33 & \underline{28.00} & 22.67 & 26.33 & 23.33 
            & 23.00 & 26.33 & 27.00 & \textbf{38.67}
            & 41.67 & 54.67 \\[0.5ex]
        \cdashline{2-15}
        \\[-2ex]
        & \textsc{L2} & 12.50 & 95.00 
            & 13.00 & 13.00 & 14.67 & 15.00 & \underline{15.33} 
            & 13.33 & 12.67 & 13.00 & \textbf{35.67}
            & 23.00 & 25.00 \\
        \midrule
        \addlinespace[5pt]
        \multicolumn{11}{l}{\textsc{\textbf{Task 5}}: \textsc{Discordant event position identification}} \\
        \midrule
        $s_1s_2s_1s_2s_1s_2$ + x & \textsc{L1} & 14.28 & 80.00 
            & \underline{24.67} & 9.67 & 18.33 & 12.00 & 8.00 
            & 13.00 & \underline{24.67} & 15.00 & \textbf{28.00}
            & 19.00 & 18.33 \\[0.5ex]
        \cdashline{2-15}
        \\[-2ex]
        $s_1s_2s_1s_2s_1s_2s_1s_2$ + x & \textsc{L2} & 11.11 & 85.00 
            & 14.67 & 10.67 & 14.67 & 12.67 & 11.00 
            & 13.00 & \underline{15.00} & 12.33 & \textbf{21.67}
            & 14.33 & 11.67 \\
        \midrule
        $s_1s_2s_3s_1s_2s_3$ + x & \textsc{L1} & 14.28 & 90.00 
            & 18.67 & 7.00 & 17.33 & 15.00 & 10.00 
            & 17.33 & \underline{20.67} & 12.67 & \textbf{21.00}
            & 21.67 & 21.67 \\[0.5ex]
        \cdashline{2-15}
        \\[-2ex]
        $s_1s_2s_3s_1s_2s_3s_1s_2s_3$ + x & \textsc{L2} & 10.00 & 95.00 
            & 13.33 & 13.67 & \underline{14.33} & 10.67 & 14.00 
            & 14.00 & 8.33 & 8.00 & \textbf{19.00}
            & 13.33 & 11.33 \\
        \bottomrule
    \end{tabular}
    }
    \vspace{-0.5em}
    \caption{\textbf{Exact match (EM) accuracy on \benchmark.} 
    The models `Video-LLaMA2', `LongLLaVA', `LLaVA-OneVision' and `Gemini 1.5 Pro' are abbreviated as `V-LM2', 'LongLV', LV-OV' and `Gem.1.5P' respectively. \method refers to our model. 
    `Lvl' indicates the task difficulty level, and `Chance' refers to the probability of a correct random guess.
    In Task~5 rows, $s_i$ denotes the $i$-th action event of the pattern $(s_1, s_2)$ or $(s_1, s_2, s_3)$, where each $s_i$ corresponds to a distinct event. For each row, the highest value is highlighted in bold, and the second-highest is underlined among open-source VLMMs with approximately 7B parameters.}
    \label{tab:main_alltasks}
    \vspace{-0.5em}
\end{table*}

\subsection{Metrics}
\label{subsec:evaluation_metrics}
To evaluate model performance on the proposed tasks, we define four metrics. The simplest is \textit{Exact Match (EM)}, which checks if predictions exactly match ground truths (used for all tasks (1–5)). Additionally, to assess fine-grained temporal sequencing abilities in Tasks 1 and 2, we introduce three metrics: \textit{Partial Match (PM)}, \textit{LCS Match (LM)}, and \textit{Orderless Match (OM)}.

\vspace{-0.5em}
\paragraph{Partial match (PM).}  
For a more fine-grained assessment of partial prediction alignment with the ground-truth, we measure the fraction of positions that match. 
Let $M$ be the total number of questions, $N$ the number of components per sequence, $\hat{y}_i$ the model’s prediction, and $y_i$ the ground truth. 
For each component $j$ in sequence $i$:
\vspace{-0.5em}
\begin{equation}
    \text{PM} = 100\times\frac{1}{MN} \sum_{i=1}^{M} \sum_{j=1}^{N} 
    \mathbbm{1}\bigl(\hat{y}_{i,j} = y_{i,j}\bigr).
    \label{eq:pm}
\end{equation}

\vspace{-0.5em}
\noindent \textbf{Longest common subsequence match (LM).}
We measure how well the model captures the \emph{overall} structure of the sequence, allowing minor positional errors. Let $\text{LCS}(\hat{y}_i, y_i)$ denote the length of the longest common subsequence~\cite{lcs} between $\hat{y}_i$ and $y_i$:
\vspace{-0.5em}
\begin{equation}
    \text{LM} = 100\times\frac{1}{M} \sum_{i=1}^{M} 
    \frac{\text{LCS}\bigl(\hat{y}_i, y_i\bigr)}{\bigl|y_i\bigr|}.
    \label{eq:lm}
\end{equation}

\vspace{-0.5em}
\noindent \textbf{Orderless match (OM).} 
This metric evaluates whether the model identifies \emph{which} events appear in the sequence, disregarding the order:
\vspace{-0.5em}
\begin{equation}
    \text{OM} = 
    100\times\frac{1}{M} \sum_{i=1}^{M} \frac{\bigl|\hat{y}_i \cap y_i\bigr|}{\bigl|y_i\bigr|}.
    \label{eq:om}
\end{equation}

\section{\method: Multi-Event Instruction Fine-tuning with Chain-of-Thought}
\label{sec:mecot}
To address the challenges of temporal order understanding, models must explicitly learn relationships among individual events and grasp the underlying temporal progression~\cite{zhou2018temporal,guo2025trace,cheng2025tempura}. 
Motivated by this, we propose \textbf{MECoT}, combining detailed event-level \emph{instruction fine-tuning} with \emph{Chain-of-Thought} (CoT) reasoning for effective temporal understanding in videos, as illustrated in Fig.~\ref{fig:temp_inst_tune} (see details in the supplementary material).

\paragraph{Instruction fine-tuning from video descriptions.}
Existing instruction-tuning methods often simplify complex temporal sequences into general summaries (\eg, combining entering a room, sitting, and typing into \emph{working at a desk}), losing essential fine-grained details~\cite{nguyen2024video,qu2024chatvtg}.
Inspired by prior work~\cite{chen2024sharegpt4video}, we maintain detailed temporal integrity by generating segment-wise descriptions with a pre-trained VLMM (7B)~\cite{li2024llavaov}, then merging these descriptions into coherent narratives using GPT-4o-mini~\cite{gpt-4o}, as illustrated in Figure~\ref{fig:temp_inst_tune}-(a).
This merging addresses the unclear transitions that can arise when simply concatenating descriptions and produces structured temporal sequences that enable the model to better learn temporal coherence.
We create 120k enriched video descriptions, used to fine-tune a 7B LLaVA-OneVision model~\cite{li2024llavaov} on 32 frames per video for one epoch at a learning rate of 2e-7 with 4$\times$NVIDIA H100 GPUs (80G). 
The resulting fully fine-tuned model forms the foundation model $\mathcal{M}$ of \method.

\vspace{0.5em}
\noindent \textbf{Chain-of-thought reasoning.}
Although multi-event instruction fine-tuning enhances temporal understanding, \emph{explicitly} articulating the reasoning process remains crucial for recognizing events in multi-event videos~\cite{zhang2024improvevisionlanguagemodel,xu2025llavacotletvisionlanguage}. 
To this end, \method adopts a Chain-of-Thought (CoT) inference strategy~\cite{xu2025llavacotletvisionlanguage,vlm-rlaif,zhang2024direct}, as illustrated in Fig.~\ref{fig:temp_inst_tune}-(b). 
Specifically, given a multi-event video $V$, the fine-tuned model $\mathcal{M}$ first generates a chronological video context $d$ by applying a generation prompt $p_{\text{gen}}$ to $V$. 
Then, the final prediction $y$ is made by combining this generated context $d$ with a query prompt $p_{\text{query}}$, guiding the model to capture and reason about the video's temporal structure.
While we do not employ step-by-step CoT prompts during training, we allow the model at inference to take the intermediate step to reason about the sequence of events prior to answering, thereby fully leveraging the model's capability. 
Table~\ref{tab:method_ablation} shows the effect of this CoT process.

\begin{table*}[h!]
    \centering
    \renewcommand{\arraystretch}{1.3}
    \setlength{\tabcolsep}{4pt}
    \resizebox{1.0\textwidth}{!}{
    \begin{tabular}{lccccccccccccccccc}
        \toprule
        \multirow{3.5}{*}{Model} & \multirow{3.5}{*}{Size} & \multicolumn{8}{c}{\textbf{Task~1}: \textsc{Event sequencing}} & \multicolumn{8}{c}{\textbf{Task~2}: \textsc{Relative event sequencing}} \\
        \cmidrule(lr){3-10}\cmidrule(lr){11-18}
        & & \multicolumn{4}{c}{L1 ($N_e=4$)} & \multicolumn{4}{c}{L2 ($N_e=8$)} & \multicolumn{4}{c}{L1 ($N_e=4$)} & \multicolumn{4}{c}{L2 ($N_e=8$)} \\
        \cmidrule(lr){3-6}\cmidrule(lr){7-10}\cmidrule(lr){11-14}\cmidrule(lr){15-18}
        & & EM~$\uparrow$ & PM~$\uparrow$ & LM~$\uparrow$ & OM~$\uparrow$ & EM~$\uparrow$ & PM~$\uparrow$ & LM~$\uparrow$ & OM~$\uparrow$ & EM~$\uparrow$ & PM~$\uparrow$ & LM~$\uparrow$ & OM~$\uparrow$ & EM~$\uparrow$ & PM~$\uparrow$ & LM~$\uparrow$ & OM~$\uparrow$ \\
        \cmidrule(lr){1-18} 
        Chance & - & 0.00 & 5.00 & 17.18 & 20.00 & 0.00 & 5.00 & 23.63 & 40.00 & 2.94 & 5.56 & 8.24 & 8.33 & 0.98 & 5.56 & 16.28 & 19.44 \\
        Human &  - & 85.00 & 95.00 & 95.00 & 95.00 & 80.00 & 96.25 & 96.88 & 96.88 & 90.00 & 92.50 & 92.50 & 92.50 & 85.00 & 91.50 & 94.00 & 94.00 \\
        \cmidrule(lr){1-18}
        LongVA~\small{\aquablue{(Zhang et al., 2024)}} & 7B & 1.33 & 32.92 & 54.42 & 63.25 
            & 0.00 & 17.04 & 41.96 & 66.08 
            & 11.33 & 15.17 & 18.67 & 18.83
            & 4.33 & 9.62 & 19.27 & 22.10 \\
        Oryx~\small{\aquablue{(Liu et al., 2024)}} & 7B 
            & 0.00 & 22.33 & 28.75 & 29.67
            & 0.00 & 11.12 & 17.88 & 21.12
            & 0.67 & 0.83 & 0.83 & 0.83
            & 0.33 & 1.31 & 4.25 & 4.38 \\
        VILA-1.5~\small{\aquablue{(Lin et al., 2024)}} & 8B
            & 0.33 & 9.92 & 27.67 & 32.50
            & 0.00 & 6.38 & 25.42 & 42.96
            & 4.67 & 7.33 & 9.50 & 9.50
            & 1.67 & 4.41 & 10.79 & 12.33 \\
        V-LM2~\small{\aquablue{(Zhang et al., 2024)}} & 7B
            & 2.33 & 27.92 & 55.08 & 75.50	
            & 0.00 & 14.62 & 36.96 & 71.00	
            & 24.67 & 30.33 & 36.17 & 36.33
            & 3.00 & 9.42 & 19.74 & 23.93 \\
        LongLV~\small{\aquablue{(Wang et al., 2024)}} & 7B
            & 0.00 & 9.42 & 26.92 & 32.92
            & 0.00 & 6.50 & 24.38 & 42.00
            & 9.33 & 12.17 & 14.33 & 14.67
            & 2.67 & 7.33 & 14.48 & 16.81 \\
        LongVU~\small{\aquablue{(Shen et al., 2025)}} & 7B
            & 12.67 & 51.00 & 67.25 & 73.08
            & 0.00 & 28.25 & 51.04 & 61.58
            & 24.33 & 31.83 & 36.33 & 36.83
            & 5.33 & 16.18 & 27.92 & 31.77 \\
        NVILA~\small{\aquablue{(Liu et al., 2025)}} & 8B
            & 20.00 & 52.50 & 71.58 & 78.50
            & \underline{0.33} & 28.12 & 57.58 & 74.08
            & 29.00 & 37.33 & 41.67 & 42.33
            & 7.00 & 18.41 & 31.56 & 34.91 \\
        LV-OV~\small{\aquablue{(Li et al., 2024)}} & 7B & \underline{23.00} & \underline{53.00} & \underline{73.25} & \underline{81.08} & \underline{0.33} & \underline{28.88} & \underline{57.88} & \underline{79.58} & \underline{40.33} & \underline{49.17} & \underline{53.83} & \underline{56.00} & \underline{8.00} & \underline{19.45} & \underline{33.06} & \underline{38.70} \\
        \textbf{MECoT} (Ours) & 7B & \textbf{41.67} & \textbf{70.33} & \textbf{80.17} & \textbf{84.00} & \textbf{4.33} & \textbf{47.00} & \textbf{66.79} & \textbf{81.21} & \textbf{55.67} & \textbf{60.67} & \textbf{67.67} & \textbf{70.17} & \textbf{8.33} & \textbf{19.81} & \textbf{35.58} & \textbf{40.46} \\
        \cmidrule(lr){1-18} 
        GPT-4o & - & 80.00 & 93.25 & 94.42 & 94.50
            & 51.67 & 87.00 & 91.88 & 92.38
            & 83.33 & 85.33 & 85.50 & 85.50 
            & 53.67 & 63.76 & 72.06 & 72.31 \\
        Gemini 1.5 Pro & - & 78.00 & 88.50 & 90.08 & 90.25
            & 59.33 & 81.42 & 86.79 & 87.79
            & 80.67 & 82.33 & 82.50 & 82.50
            & 56.33 & 61.21 & 69.05 & 69.16 \\
        \bottomrule
    \end{tabular}
    }
    \vspace{-0.5em}
    \caption{\textbf{Performance on event sequencing tasks (Task 1 and Task 2).} 
    We evaluate VLMMs at two difficulty levels (\textbf{L1}, \textbf{L2}) using four metrics: EM, PM, LM, and OM.
    The models `Video-LLaVA', `Video-LLaMA2', `LongLLaVA’, and `LLaVA-OneVision' are abbreviated as `V-LV', `V-LM2', `LongLV', and `LV-OV', respectively, while \method refers to our model.
    `Chance' represents the probability of a correct random guess.  
    Highest and second-highest scores among open-source VLMMs are marked in bold and underlined, respectively.}
    \label{tab:main_sequencing}
\end{table*}


\vspace{-0.5em}
\section{Experiments}
\label{sec:temp_inst}
We evaluate our proposed \method alongside various open-source VLMMs ($\sim$7B; \cite{lin2023vila,wang2024longllavascalingmultimodalllms,zhang2024longva,damonlpsg2024videollama2,liu2024oryx,li2024llavaov, nvila}) and leading proprietary models (GPT-4o, Gemini 1.5 Pro) on our \benchmark ($\S$\ref{subsec:vector_bench_res}) and existing video benchmarks ($\S$\ref{subsec:existing_bench_res}), followed by detailed analyses ($\S$\ref{subsec:detailed_anal}). 
Unless specified, we use 32 input frames per video. 
All evaluated VLMMs achieve high single-event recognition accuracy ($> 90\%;$ see the supplementary material), providing strong yet tractable baselines for assessing multi-event temporal understanding.

\subsection{Evaluation on~\benchmark Benchmark}
\label{subsec:vector_bench_res}
\paragraph{Temporal order of `events'.}  
Tables~\ref{tab:main_alltasks} (Tasks1--3) and~\ref{tab:main_sequencing} (Tasks1--2) present the evaluation of event-order understanding in \benchmark at two difficulty levels (\textbf{L1}, \textbf{L2}). 
Specifically, Table~\ref{tab:main_alltasks} reports EM scores for Tasks1--3, while Table~\ref{tab:main_sequencing} provides additional sequencing metrics.

Overall, temporal understanding of multiple events remains challenging for VLMMs. 
Human annotators achieve high EM scores (80--100\%), confirming clear task design. 
Proprietary models (GPT-4o, Gemini 1.5 Pro) perform close to humans at L1, but open-source VLMMs score significantly lower.
In particular, all models show notable performance declines from L1 to L2; for example, GPT-4o and Gemini drop from approximately 75--83\% EM at L1 to 50--60\% at L2. 
Open-source VLMMs approach chance-level accuracy on longer (L2) sequences, highlighting their struggles in maintaining long-range temporal reasoning.

Our proposed \method outperforms other open-source VLMMs, achieving overall improvement in EM scores across Tasks1--3 (Tab.~\ref{tab:main_alltasks}) and delivering consistent gains in all sequencing metrics (EM, PM, LM, OM; Tab.~\ref{tab:main_sequencing}). 
In particular, for sequencing tasks (Tasks~1 and 2), improvements are more evident in order-sensitive metrics (EM, PM, LM) compared to the order-agnostic metric (OM), suggesting enhanced temporal-order comprehension and validating the impact of our event-level fine-tuning and structured chain-of-thought inference in modeling temporal relationships.

\vspace{-1.0em}
\paragraph{Temporal order of `pattern'.} 
Table~\ref{tab:main_alltasks} summarizes EM scores for Tasks~4 and 5, which involve identifying a single outlier event—either belonging to a different semantic group or disrupting a repeated pattern. 
Humans achieve consistently high EM scores (80--100\%), underscoring the intuitive nature of detecting anomalies. 
VLMMs' performance varies significantly based on sequence length, \ie, from L1 to L2; even proprietary models (GPT-4o, Gemini 1.5 Pro) remain notably below human accuracy and VLMMs including proprietary models exhibit substantial declines at L2. 
Increased repetitions of pattern-conforming events at L2 complicate outlier detection, driving VLMM performance toward chance and highlighting their difficulty with extended temporal contexts.

Our proposed \method surpasses comparable 7B baselines on all tasks across both difficulty levels.
We conjecture these improvements result from enhanced reasoning about complex temporal patterns, enabled by explicit multi-event instruction fine-tuning combined with structured CoT inference.
Nevertheless, effectively capturing such temporal structures remains challenging for VLMMs compared to human performance.

\begin{table}[t]
    \centering
    \resizebox{0.85\linewidth}{!}{
    \begin{tabular}{lcc}
        \toprule
        Benchmarks & LV-OV~\cite{li2024llavaov} & \textbf{\method} (Ours) \\
        \cmidrule(lr){1-1} \cmidrule(lr){2-3}
        EgoSchema & 60.10 & \textbf{62.25} \\
        MVBench & 56.70 & \textbf{57.56} \\
        LongVideoBench & 51.91 & \textbf{52.13} \\
        VideoVista & 72.99 & \textbf{74.23} \\
        MLVU & 63.17 & \textbf{63.28} \\
        TVBench & 45.29 & \textbf{45.42} \\
        TemporalBench & 60.46 & \textbf{60.79} \\
        TempCompass & 64.11 & \textbf{66.53} \\
        VidComposition & 53.99 & \textbf{55.57} \\
        \bottomrule
    \end{tabular}
    }
    \caption{\textbf{Comparison of \method with LV-OV on general benchmarks.} 
    We compare our proposed \method with its baseline model, \ie, LLaVA-OneVision (LV-OV)~\cite{li2024llavaov}. \method outperforms its baseline on various general video understanding benchmarks, showing its effectiveness on general video understanding.
    }
    \label{tab:prevbenchmark_generalization}
\end{table}

\subsection{Evaluation on Existing Video Benchmarks}
\label{subsec:existing_bench_res}
To evaluate the generalizability of \method beyond our proposed \benchmark benchmark, we compare it against its baseline~\cite{li2024llavaov} across various established video benchmarks, as summarized in Tab.~\ref{tab:prevbenchmark_generalization}. 
\method consistently outperforms its baseline, demonstrating improved general video understanding. 
These gains suggest that explicitly modeling complex temporal relationships through detailed multi-event instruction fine-tuning and structured CoT inference helps the model better capture temporal structures, which are essential for a wide range of video comprehension tasks.

\subsection{Comparison with Video Reasoning Methods}
\label{subsec:comparison_video_reasoning}
Moreover, to evaluate the effectiveness of our proposed \method, we compare it against existing video reasoning approaches. 
Specifically, we apply CoT-based methods~\cite{wang2024videoagent, zhang2023llovi} to the baseline model, LLaVA-OneVision (LV-OV) 7B~\cite{li2024llavaov}, and evaluate performance on EgoSchema~\cite{mangalam2023egoschema}, a benchmark requiring long-context temporal understanding in videos.
While prior methods~\cite{wang2024videoagent, zhang2023llovi} demonstrate competitive results compared to the baseline, our \method achieves state-of-the-art performance, as shown in Figure~\ref{tab:comparison_video_reasoning}. 
We conjecture this improvement stems from our event-level fine-tuning combined with structured CoT prompting, which explicitly guides the model through temporal event reasoning. 
Consequently, MECoT more effectively captures complex temporal dependencies, demonstrating clear advantages over existing CoT-based approaches.

\begin{table}
    \centering
    \resizebox{0.7\linewidth}{!}{
    \begin{tabular}{lc}
        \toprule
        Methods & Accuracy \\
        \cmidrule(lr){1-1} \cmidrule(lr){2-2}
        Baseline (LV-OV~\cite{li2024llavaov}) & 60.10 \\
        Baseline + VideoAgent~\cite{wang2024videoagent} & 61.10 \\
        Baseline + LLoVI~\cite{zhang2023llovi} & 61.66 \\
        \textbf{\method} (Ours) & \textbf{62.25} \\
        \bottomrule
    \end{tabular}
    }
    \caption{\textbf{Comparison of \method with video reasoning methods on EgoSchema.} 
    MECoT outperforms existing video reasoning methods on the general video understanding benchmark, EgoSchema~\cite{mangalam2023egoschema}, demonstrating its effective temporal understanding in videos.
    }
    \label{tab:comparison_video_reasoning}
\end{table}

\subsection{Detailed Analysis}
\label{subsec:detailed_anal}
For detailed analysis, we evaluate the \emph{event sequencing task} (Task 1), which directly measures VLMMs' ability to understand multi-event order.

\vspace{-1.0em}
\paragraph{Effect of proposed components in~\method.}
Table \ref{tab:method_ablation} evaluates two key components: \emph{multi-event instruction fine-tuning} (IFT) and \emph{chain-of-thought} (CoT). 
We compare four settings: baseline (LV-OV), CoT alone, IFT alone, and their combination (\method). 
CoT alone yields only marginal improvements, showing limited effectiveness without targeted event-level training. 
In contrast, IFT notably boosts performance, underscoring the value of explicit instruction fine-tuning for learning event sequences. 
Combining CoT with IFT (\method) further enhances performance, indicating complementary benefits from structured reasoning. 
OM scores remain stable, but significant improvements in order-sensitive metrics (EM, PM, LM) demonstrate enhanced temporal reasoning capability.

\begin{table}[t]
    \setlength{\tabcolsep}{3.2pt}
    \centering
    \resizebox{1\linewidth}{!}{
    \begin{tabular}{lcccccccc}
        \toprule
        \multirow{4}{*}{Method} 
        &\multicolumn{8}{c}{\textbf{Task~1}: \textsc{Event Sequencing}} \\
        \cmidrule(lr){2-9}
        & \multicolumn{4}{c}{\textbf{L1} ($N_e=4$)} & \multicolumn{4}{c}{\textbf{L2} ($N_e=8$)} \\
        \cmidrule(lr){2-5}\cmidrule(lr){6-9}
        & EM~$\uparrow$ & PM~$\uparrow$ & LM~$\uparrow$ & OM~$\uparrow$ & EM~$\uparrow$ & PM~$\uparrow$ & LM~$\uparrow$ & OM~$\uparrow$ \\
        \midrule
        LV-OV
            & 23.00 & 53.00 & 73.25 & 81.08 
            & 0.33 & 28.88 & 57.88 & 79.58 \\
            ~ + CoT
            & 23.67 & 48.92 & 72.25 & 81.42
            & 0.67 & 30.33 & 55.62 & 75.54 \\
            ~ + IFT
            & 35.33 & 62.50 & 77.92 & \textbf{84.33}
            & 1.33 & 33.62 & 61.00 & \textbf{81.71} \\
            
            ~ + IFT + CoT
            & \textbf{41.67} & \textbf{70.33} & \textbf{80.17} & 84.00 
            & \textbf{4.33} & \textbf{47.00} & \textbf{66.79} & 81.21  \\
        \bottomrule
    \end{tabular}
    }
    \caption{\textbf{Impact of \method components on event sequencing task at varying difficulties.} `IFT' and `CoT' denote instruction fine-tuning and chain-of-thought, respectively. 
    Combining both components (\ie, \method in fourth row) significantly boosts performance.
    } 
    \label{tab:method_ablation}
\end{table}

\begin{figure}[t]
    \centering
    \includegraphics[width=1\linewidth]{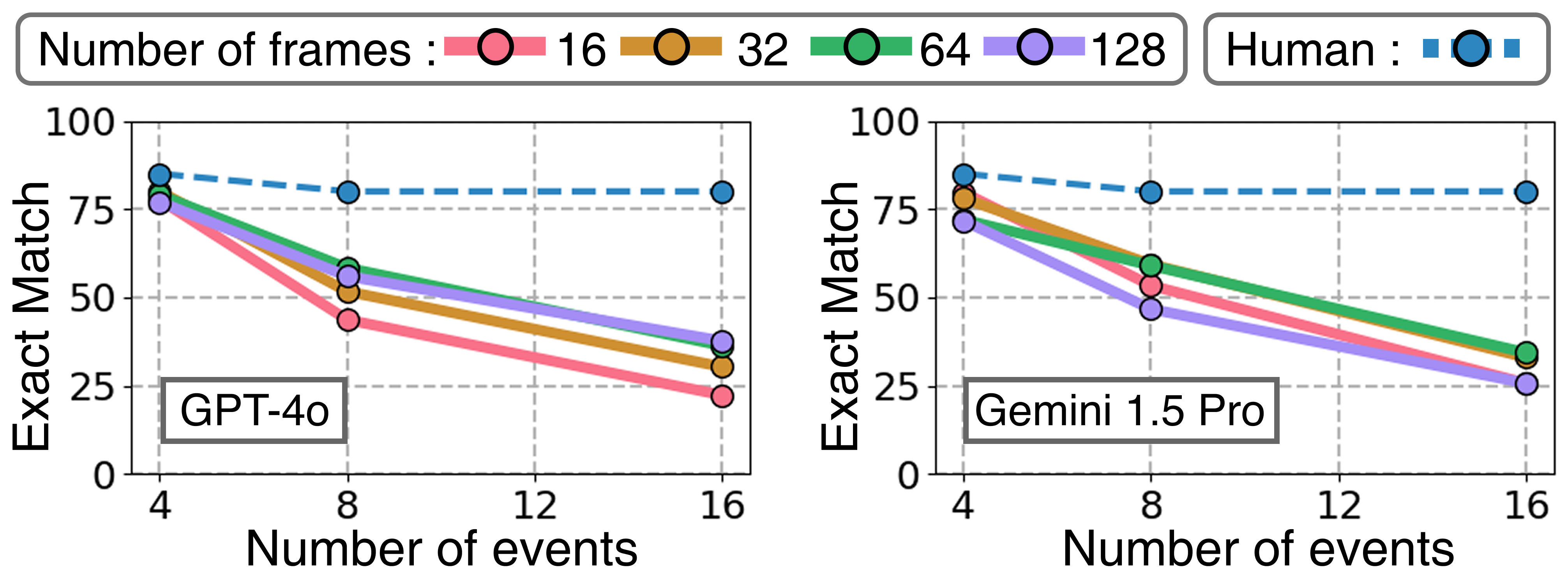}
    \caption{\textbf{EM on event sequencing (Task 1) with varying event counts and input frames.}
    We evaluate two proprietary VLMMs across different event and frame counts (indicated by legend). Even SoTA proprietary VLMMs exhibit substantial performance degradation as the number of events increases, while human performance remains stable. 
    Notably, increasing input frames does not consistently enhance model performance.}
    \label{fig:in_depth}
\end{figure}

\vspace{-1.0em}
\paragraph{The number of events and input frames.} 
Figure~\ref{fig:in_depth} shows how EM accuracy in event sequencing  task (Task 1) varies with event count (x-axis) and frame number (lines) for GPT-4o~\cite{gpt-4o} and Gemini 1.5 Pro~\cite{gemini}. 
As events increase from 8 to 16, performance of even proprietary models sharply declines, lagging \emph{far behind} humans and highlighting their limitations in longer temporal contexts. 
Further, adding more frames beyond a certain threshold offers no improvement and can even degrade performance. 
These observations suggest that simply providing more visual information is insufficient; instead, models must explicitly learn meaningful temporal relationships among events to achieve robust comprehension of longer videos.

\section{Conclusion}
\label{sec:conclusion}
We empirically highlight a key limitation: existing benchmarks inadequately assess temporal understanding, as VLMMs perform similarly even on frame-shuffled videos. 
Evaluations reveal significant challenges for VLMMs in comprehending temporal relationships among events in videos. 
To address this, we introduce \benchmark, specifically designed to measure event-order understanding in videos. 
Further, we propose \method, a simple yet effective method combining instruction fine-tuning with CoT reasoning to enhance temporal reasoning explicitly. 
Experiments confirm that \method improves performance on both \benchmark and existing video understanding benchmarks.

\section*{Acknowledgement}
This work was partly supported by the IITP grants (RS-2022-II220077, RS-2022-II220113, RS-2022-II220959, RS-2022-II220871, RS-2021-II211343 (SNU AI), RS-2025-25442338 (AI Star Fellowship-SNU)) funded by the Korea government (MSIT), a grant (No. RS-2025-25453780) funded by MOTIE, a grant of Korean ARPA-H Project through the Korea Health Industry Development Institute (KHIDI) funded by the Ministry of Health \& Welfare, Republic of Korea (RS-2025-25424639), Artificial Intelligence Industrial Convergence Cluster Development Project funded by the Ministry of Science and ICT(MSIT, Korea) \& Gwangju Metropolitan City, and the BK21 FOUR program, SNU in 2025.

{
    \small
    \bibliographystyle{ieeenat_fullname}
    \bibliography{main}
}

\clearpage
\setcounter{page}{1}
\maketitlesupplementary

\newcommand{\bmp}[1]{\textcolor{blue}{#1}} 
\newcommand{\rmp}[1]{\textcolor{red}{#1}} 
\newcommand{\gmp}[1]{\textcolor{green}{#1}} 
\newcommand{\tmp}[1]{\textcolor{teal}{#1}} 

\small \noindent \textbf{Note}: We use \bmp{blue} color to refer to figures, tables, section numbers, and citations \textbf{in the main paper} (\eg, ~[\bmp{10}]).
We use \cblue{steel blue} color to refer to figures, tables, section numbers, and citations \textbf{in this supplementary material}.

\begin{figure*}[t]
    \centering
    \includegraphics[width=1.0\textwidth]{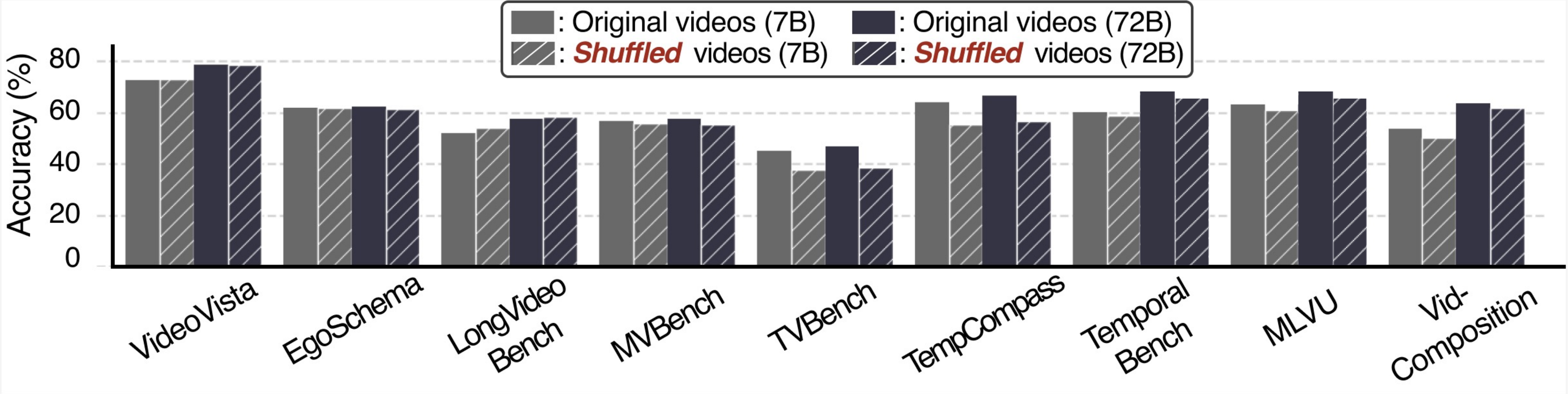}
    \vspace{-0.9em}
    \caption{\textbf{Accuracy of VLMMs on original vs. frame-shuffled videos across different benchmarks.} We evaluate the accuracy of LLaVA-OV 7B and 72B~[\bmp{15}] across nine video understanding benchmarks. We observe that there's no significant performance difference between evaluations using original videos and frame-shuffled videos across all benchmarks.
    }
    \label{fig:supp_shuffled_video_prevbench_acc}
\end{figure*}

\begin{table*}[t]
\centering
\resizebox{\textwidth}{!}{
\begin{tabular}{c cccccccc}
\toprule
\multirow{2.5}{*}{Models} & \multicolumn{8}{c}{$\rho$} \\
\cmidrule(lr){2-9}
 & MVBench & EgoSchema & LongVideoBench & TempCompass & TVBench & VideoVista & TemporalBench & MLVU \\
\midrule
LLaVA-OV (7B) & 98.59 & 99.36 & 104.4 & 86.35 & 83.68 & 100.1 & 97.25 & 96.56 \\
LLaVA-OV (72B) & 96.01 & 98.77 & 101.4 & 85.57 & 82.71 & 100.1 & 96.22 & 95.58 \\
\bottomrule
\end{tabular}}
\vspace{-0.5em}
\caption{\textbf{Performance ratio ($\rho$) of VLMMs: Comparison of model performance on original \vs frame-shuffled videos across different benchmarks.} We measure performance ratio ($\rho$) of LLaVA-OV 7B and 72B on seven video understanding benchmarks. $\rho$ represents the ratio of benchmark performance on frame-shuffled videos compared to their performance on original videos. We observe no significant performance difference between evaluations using original videos and frame-shuffled videos across all benchmarks.}
\label{tab:supp_prevbench_shuffled_video}
\end{table*}

\begin{table*}[ht]
\centering
\resizebox{\textwidth}{!}{
\begin{tabular}{lcccccccc}
\toprule
\multirow{2.5}{*}{Benchmark}
& \multirow{2.5}{*}{\shortstack{Answer\\Type}}
 &  &  
 & \multirow{2.5}{*}{\shortstack{Duration\\(Avg./Max)}} & 
 & \multirow{2.5}{*}{\shortstack{Difficulty\\Level}}
 & \multirow{2.5}{*}{\shortstack{Pattern\\Level}}
 & \\
 &  & \#Events & \#Videos & & \#QA & & & Main Goal \\
\midrule
VideoVista & MCQ & 2-10 & 3,402 & 131s / 919s & 25,000 & O & X & long video understanding \\
EgoSchema & MCQ & - & 5,031 & 180s / 180s & 5,031 & X & X & long egocentric video understanding \\
LongVideoBench & MCQ & - & 3,763 & 477s / 60m & 6,678 & O & X & long video understanding \\
MVBench & MCQ & - & 3,655 & 18s / 176s & 4,000 & O & X & comprehensive  video understanding \\
TVBench & MCQ & 2-13 & 2,217 & 21s / 116s &  2,525 & X & X & video temporal understanding \\
TempCompass & MCQ/Open & 1-2 & 410 & 11s / 35s & 7,540 & X & X & short video temporal understanding \\
TemporalBench & Binary & 4-8 & 3,753 & 46s / 20m & 9,867 & O & X & video caption matching \\
MLVU & MCQ/Open & 4 & 1,112 & 755s / 133m & 2,174 & X & X & long video understanding \\
VidComposition & MCQ & 3-8 & 982 & 26s / 139s & 1,706 & X & X & compiled video understanding \\
\cmidrule{1-9}
VECTOR & List/MCQ & 4-9 & 31,200 & 64s / 100s
 & 4,800 & O & O & temporal order understanding \\
\bottomrule
\end{tabular}}
\vspace{-0.5em}
\caption{\textbf{Statistics of \benchmark compared with previous benchmarks.} Unlike other benchmarks, \benchmark considers pattern-level order. For comprehensive benchmarks, we count \# Events for the benchmarks with event ordering task.}
\label{tab:supp_vector_statistics}
\end{table*}

\section{Further Discussion on the Impact of Shuffled Video Frames in Previous Video Understanding Benchmarks}
\label{app:further_discussions}
The ability to comprehend event sequences in videos is crucial for VLMMs to achieve human-like understanding of real-world visual scenarios. 
Recent benchmarks evaluate the temporal understanding of VLMMs through question-answering tasks~[\bmp{11, 17, 18, 19, 24, 29, 38, 44}], primarily focusing on temporal relationships and causal reasoning.

However, our preliminary experiments reveal that state-of-the-art (SoTA) VLMMs~[\bmp{15}] demonstrate notably strong performance even when video frames are randomly shuffled (Fig.\bmp{1}-(a)). 
This indicates that existing benchmarks often enable correct answers without the necessity of genuine temporal comprehension.
In particular, as illustrated in Fig.\bmp{1}, both 7B and 72B variants of LLaVA-One-Vision~[\bmp{15}] exhibit similar robustness when temporal frame order is disrupted, as supported by Fig.\ref{fig:supp_shuffled_video_prevbench_acc}. 
Specifically, Tab.\ref{tab:supp_prevbench_shuffled_video} shows that across nine benchmarks, both model scales retain over 82\% of their original accuracy ($\rho$) even with temporally shuffled input. 
This consistency across model scales further supports the observation that current benchmarks inadequately assess genuine temporal understanding capabilities.
We hypothesize that models primarily rely on their \emph{prior knowledge} of common event scenarios, inferring plausible contexts from isolated frames rather than explicitly analyzing temporal relationships depicted in the videos~[\bmp{7, 47}]. 
Consequently, VLMMs may bypass true temporal reasoning by using common-sense shortcuts.




\section{Comprehensive Comparison with Existing Video Understanding Benchmarks}
\label{app:supp_review_prior_benchmarks}
We compare detailed properties of existing benchmarks and highlight differences with \benchmark.  
Table~\ref{tab:supp_vector_statistics} provides a statistical comparison of our proposed \benchmark benchmark against various existing video understanding benchmarks. 
Unlike most prior benchmarks, \benchmark explicitly evaluates temporal order understanding at both event and pattern levels. It includes 31,200 videos covering 4,800 questions, each video containing 4–9 clearly defined events. 
The videos have an average duration of 64 seconds, ranging from 4 to 100 seconds, providing concise yet temporally structured sequences. 
In contrast, previous benchmarks primarily rely on multiple-choice (MCQ) or binary answers and tend to evaluate only event-level comprehension. 
\benchmark uniquely integrates both event and higher-level temporal reasoning tasks, specifically targeting temporal order understanding, thus offering a comprehensive diagnostic tool for video multimodal models.

Detailed descriptions of each general video benchmark and further comparisons to the proposed \benchmark benchmark are provided below.

\paragraph{TempCompass.}
TempCompass~[\bmp{24}] evaluates temporal perception in Video Large Multimodal Models (VLMMs) across aspects such as action, speed, direction, attribute change, and event order.
It introduces conflicting video pairs—where static content remains unchanged while temporal aspects vary—to mitigate single-frame bias and reliance on language priors.
TempCompass also expands evaluation formats beyond multiple-choice QA, incorporating caption matching and generation to assess models' generalization across different response styles.
Although TempCompass covers diverse temporal phenomena, it is focused on short videos that contains less than two events. Also, many of its questions could still be answered by analyzing individual or pairwise frames rather than reasoning over multiple successive events.

\paragraph{MLVU.} 
MLVU~[\bmp{57}] evaluates long video understanding across various tasks and genres. 
One of MLVU’s key tasks involves embedding short `probe' events into a lengthy background video, necessitating that models first locate these target clips amid large amounts of noisy frames, and then interpret their order.
This setup does incorporate an element of temporal reasoning; however, it often centers on searching through excessive context to spot the relevant information, rather than systematically examining the temporal sequence of every event in the video.
In contrast, \benchmark focuses specifically on multi‐event \emph{ordering}, using intentionally concatenated short clips to create a series of distinct events.
Furthermore, \benchmark not only evaluates models on full-sequence ordering but also introduces various tasks such as sub-sequence ordering with detailed metrics over exact match that improves understanding of model's failure cases, enabling more comprehensive assessment for global temporal understanding.

\paragraph{TemporalBench.}
TemporalBench~[\bmp{4}] is a benchmark for evaluating fine-grained temporal understanding in videos. 
It contatins approximately 10K QA pairs derived from human annotations, covering temporal reasoning skilss such as action frequency, motion magnitude, and event order. 
The benchmark supports both video QA and captioning across short and long videos, thereby offering a comprehensive testbed for assessing multimodal video models.  
The key difference between TemporalBench and our \benchmark lies in their evaluation scope.
TemporalBench primarily measures whether models capture temporal dynamics across entire individual video clips, relying on a binary caption selection and a single, generic caption generation (``Please generate a caption for the following video'').  
In contrast, \benchmark introduces a multi-layered evaluation framework with diverse tasks explicitly designed to assess a model’s understanding of temporal relationships among multiple events and patterns.

\paragraph{EgoSchema.} 
EgoSchema~[\bmp{29}] is a diagnostic benchmark designed to assess long-form video-language understanding in VLMMs. 
It introduces the notion of temporal certificate length, quantifying the intrinsic temporal difficulty of video understanding tasks. 
It consists of over 5000 multiple-choice QAs curated to require reasoning over long, diverse egocentric videos, with questions demanding extended temporal comprehension beyond short-term cues. 
While EgoSchema focuses on long-form video comprehension and memory, often posing questions about events that are inherently difficult to segment and thus more easily answered using prior knowledge, \benchmark is specifically designed to prevent reliance on such priors, ensuring a more effective evaluation of temporal understanding.

\paragraph{MVBench.}
MVBench~[\bmp{17}] evaluates temporal perception and reasoning in video large multimodal models (VLMMs) by adapting static image tasks into 20 video-based tasks (\eg, action sequence, moving direction).  
It covers action recognition, object tracking, scene transitions, and reasoning, requiring models to process temporal changes rather than single frames.  
Using 11 public datasets, MVBench automatically generates multiple-choice QAs with LLMs, ensuring broad and diverse evaluation.  
While MVBench assesses general video understanding, \benchmark focuses on event sequencing and temporal order comprehension.  
Unlike multiple-choice QAs, \benchmark requires structured event predictions, making it more resistant to shortcuts and prior-knowledge reliance.  

\paragraph{VideoVista.}
VideoVista~[\bmp{19}] is a large-scale, versatile benchmark designed to comprehensively evaluate video understanding and reasoning capabilities of VLMMs. 
It consists of 25,000 questions spanning multiple categories, covering both short and long videos (ranging from a few seconds to over 10 minutes). 
VideoVista includes various tasks related to temporal reasoning, such as event sequence prediction and action localization. However, its primary focus is on diverse aspects of video comprehension rather than specifically evaluating the fine-grained understanding of multi-event temporal order. In contrast, \benchmark specifically targets a model’s ability to comprehend and reason about event sequences within videos.

\paragraph{TVBench.}
TVBench~[\bmp{11}] is a multiple-choice video-language benchmark designed to evaluate the temporal understanding capabilities of VLMMs. 
It focuses on assessing event sequencing, temporal localization, and the ability to distinguish fine-grained temporal relationships in video data. 
Unlike many existing video QA benchmarks that contain spatial and textual biases, TVBench ensures that solving the tasks requires genuine temporal understanding by carefully designing questions and answer choices. 
The benchmark includes a diverse set of tasks, such as action sequencing, object movement tracking, and scene transitions, sourced from various real-world and synthetic datasets. 
Compared to TVBench, which primarily serves as a diagnostic benchmark for analyzing the limitations of current VLMMs in handling temporal dependencies, VECTOR introduces a novel dataset specifically designed to evaluate temporal order comprehension.


\paragraph{LongVideoBench.}
LongVideoBench~[\bmp{44}] is a benchmark designed to evaluate long-context multimodal understanding in video-language models. 
It assesses whether models can retrieve and reason about specific details within hour-long videos using referring reasoning tasks, which require models to locate relevant moments and answer complex multimodal questions. 
Unlike LongVideoBench, which focuses on long video comprehension and multimodal reasoning, \benchmark is specifically designed to assess event sequencing and temporal order understanding in shorter multi-event videos. 
\benchmark evaluates whether models can correctly order events based on temporal cues, while LongVideoBench includes some temporal reasoning tasks that can often be solved using prior knowledge rather than requiring models to derive the actual event sequence from visual evidence.

\paragraph{VidComposition.}
VidComposition~[\bmp{38}] is a benchmark evaluating fine-grained video composition understanding of Multimodal Large Language Models (MLLMs). 
It comprises 982 videos and 1,706 multiple-choice questions, emphasizing compositional elements like camera movement, angles, shot size, narrative structure, character actions, and emotions. 
Compared to other general video benchmarks, VidComposition focuses on nuanced interpretation of complex visual contexts. Evaluations of 33 MLLMs highlight substantial gaps compared to human performance, revealing areas for improvement.

Unlike VidComposition, which assesses visual composition and narrative interpretation, \benchmark specifically targets models' ability to explicitly reason about temporal order across discrete video events. While VidComposition emphasizes cinematographic and emotional context, \benchmark uses controlled synthetic videos to deliberately disrupt common-sense priors, explicitly evaluating temporal reasoning independent of prior knowledge.



\section{Experimental Details For Diagnosing the Prior-Knowledge Shortcut}
\label{app:details_diagnosing_prior_knowledge_shortcut}
To investigate whether VLMMs rely more on prior knowledge than true temporal understanding, we conduct an empirical study using two densely captioned video datasets: MECD~[\bmp{7}] and HiREST~[\bmp{49}].  
Both datasets provide fine-grained captions aligned with specific video segments, enabling precise temporal analysis.  
We preprocess these datasets to construct an evaluation set that assesses whether models can accurately infer event order based on visual content rather than relying on common-sense priors.  

To construct the evaluation set, we select non-overlapping captions from both datasets to ensure that each caption represents a distinct event.  
To minimize the possibility of models inferring event order without relying on visual content, we refine the captions using GPT-4o.  
Specifically, we eliminate references such as ``that man'' or ``then'', which could implicitly encode temporal relationships between events.  

Using this refined dataset, we conduct an evaluation where models are presented with a set of event descriptions labeled with option letters (A, B, C, D) and tasked with reordering them into their correct chronological sequence based on the video input.  
The dataset consists of 534 samples from MECD and 296 samples from HiREST, where each sample contains multiple captions describing different events in the video.  
Each sample includes 3, 4, 5, or 6 distinct events.


\section{Building \benchmark Benchmark}
\label{app:building_vector_benchmark}
\subsection{Data Construction for Understanding Temporal Order of Events}
We constructed \benchmark rule-based to allow the scalable evaluation.
For event-level temporal order understanding tasks, we ensure that each multi-event video $V$ comprises distinct events.
We first sample subset of event categories $C^e$ which is feed to the model to list the recognized events from all 700 Kinetics action classes~[\bmp{37}].
We then randomly sample $N_e$ distinct events and their corresponding videos from the validation split of Kinetics-700 dataset.
We made sure that all $N_e$ event types are distinct to avoid ordering confusion.
We define two difficulty levels by varying the sequence length $N_e$, $N_e=4$ for L1 and $N_e=8$ for L2.
Resulting event sequencing task consists of 3k questions across three tasks, with each task evaluated at two task difficulty levels (L1 and L2):
\begin{itemize}
    \item Event sequencing: 0.6k questions (0.3k per difficulty level), using 2.5k videos.
    \item Relative event sequencing: 0.6k questions (0.3k per difficulty level), using 2.5k videos.
    \item Event position identification: 1.8k questions (0.6k for each single, double, and triple event identification), using 8.1k videos.
\end{itemize}

\subsection{Data Construction for Understanding Temporal Order of Patterns}
For pattern-level temporal order understanding tasks, we carefully construct multi-event videos $V$ by selecting events that align with specific task requirements. 
We sample events from our semantic groups $\mathcal G$, which is a set of semantic superclasses of Kinetics action classes~[\bmp{37}], using videos from the validation split of Kinetics-700 dataset. 
The dataset consists of 1.8k questions across two reasoning tasks:

\begin{itemize}
    \item Discordant semantic-group position identification: 0.6k questions (0.3k per difficulty level), using 3.1k videos.
    \item Discordant event position identification: 1.2k questions (0.3k for each of four event patterns), using 8.6k videos.
\end{itemize}

\paragraph{Construction for semantic group $\mathcal G$.}
We organize the 700 action classes from the Kinetics dataset into semantic groups $\mathcal{G}$ through an iterative refinement process involving human-LLM collaboration (Claude 3.5~[\bmp{2}]).  
Initially, Claude 3.5 generated 50 preliminary groups, followed by three rounds of refinement:  
(1) expert review by a vision-language researcher,  
(2) reorganization based on expert judgment, and  
(3) LLM validation and refinement of the revised groupings.  

\begin{figure*}[t]
    \centering
    \includegraphics[width=0.84\linewidth]{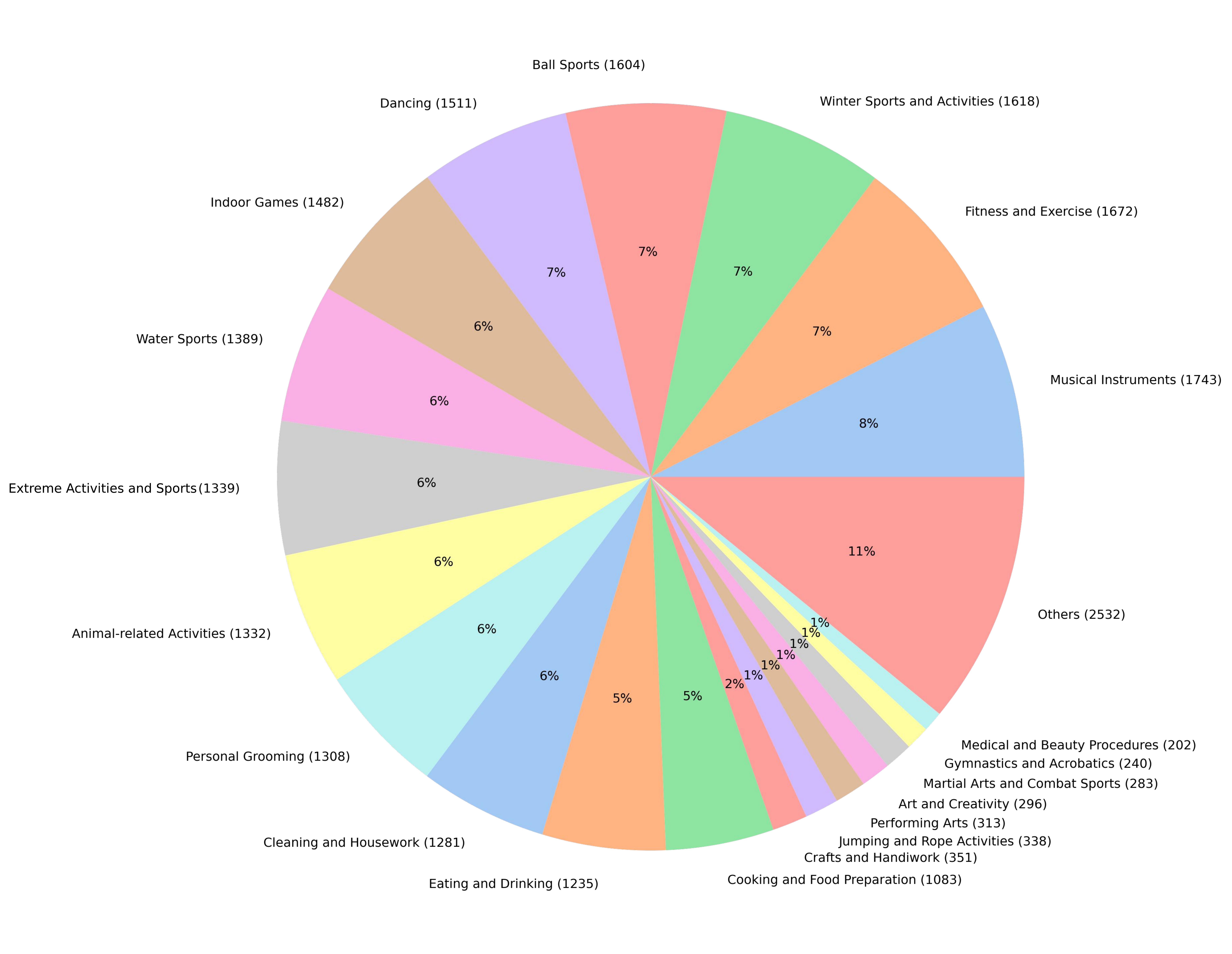}
    \caption{\textbf{Top 20 subset of semantic group $\mathcal G$.} We visualize the top 20 semantic groups for visualization clarity, used in constructing multi-event videos. These groups are derived from Kinetics-700 classes~[\bmp{37}] through an iterative refinement process. Initially, we use Claude 3.5~[\bmp{2}] to create 50 preliminary groups, followed by three rounds of revision involving expert review by a vision-language researcher and subsequent refinement using Claude 3.5. The values in parentheses indicate the number of video instances within each event group.
    }
    \label{fig:supp_top20_groups}
    \vspace{-1em}
\end{figure*}

Rare cases that do not fit into clear categories are consolidated into a `Miscellaneous Activities' group, which is excluded from our data construction to maintain clarity in group separation.  
Through this three-iteration refinement, we categorize the 700 classes into 50 distinct semantic groups.  
Figure~\ref{fig:supp_top20_groups} visualizes the top 20 groups for clarity, with `Others' representing the aggregate of the remaining 30.  
The values in parentheses indicate the number of video instances within each group.

\paragraph{Discordant semantic-group position identification.}
We randomly choose two semantic groups from $\mathcal G$: a dominant group $g^d$ and a discordant group. 
For the dominant group $g^d$, we sample $N_e - 1$ different events and their corresponding videos from Kinetics validation split dataset. 
We then sample one event from the discordant group and position it at the $k$-th position in the sequence of events, as illustrated in the Task~4 of Fig.~\ref{fig:supp_prompt_pattern}.
The VLMM's task is to identify the position $k$ where the event's semantic group $g_{e_k}$ differs from the dominant group $g^d$, while all other positions contain events that belong to the dominant group ($g_{e_i} = g^d$ for $i \ne k$).
For this discordant semantic-group position identification task, we collected 300 multi-event videos with corresponding question-answer pairs.

\paragraph{Discordant event position identification.}
For discordant event position identification, we first define four types of event patterns: $s_1s_2s_1s_2s_1s_2$, $s_1s_2s_1s_2s_1s_2s_1s_2$, $s_1s_2s_3s_1s_2s_3$, and $s_1s_2s_3s_1s_2s_3s_1s_2s_3$, where each $s_i$ represents events from different classes.
For each pattern, the task is to identify a position where a randomly injected discordant event $x$ ($x \ne s_i$) disrupts the repeating pattern. 
We collected 300 multi-event videos with corresponding question-answer pairs for each pattern type.

\subsection{Detailed Evaluation Prompts for Each Task}
\begin{figure*}[ht]
    \includegraphics[width=1\linewidth]{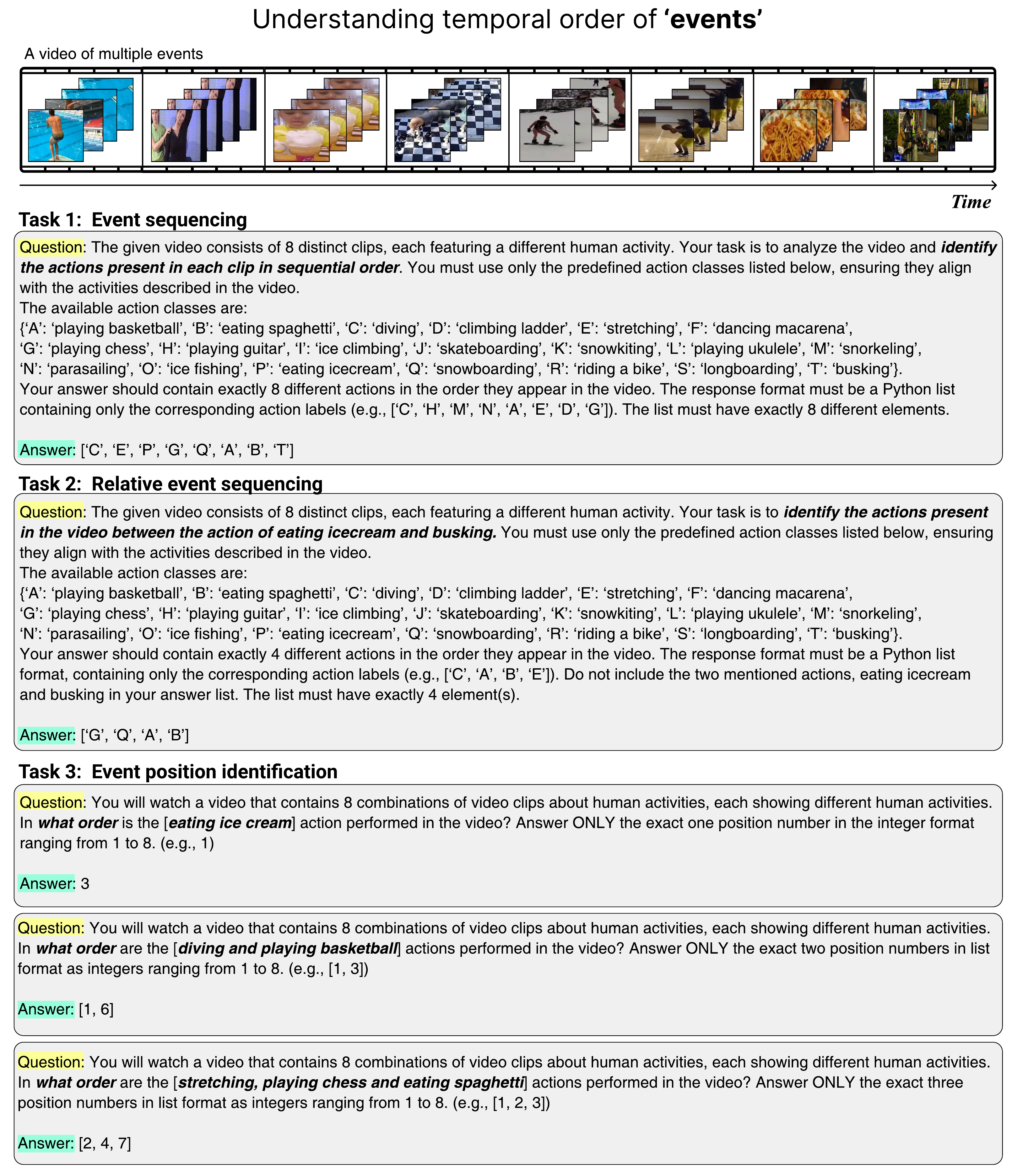}
    \caption{\textbf{Prompt details employed for understanding temporal order of events.} We decompose the event-level temporal order understanding task into three sub-tasks: event sequencing task, relative event sequencing task and event position identification task. We divide event position identification task into three variations (1 to 3) according to the number of events to be identified.
    }
    \vspace{-1em}
    \label{fig:supp_prompt_event}
\end{figure*}

\begin{figure*}[ht]
    \includegraphics[width=1\linewidth]{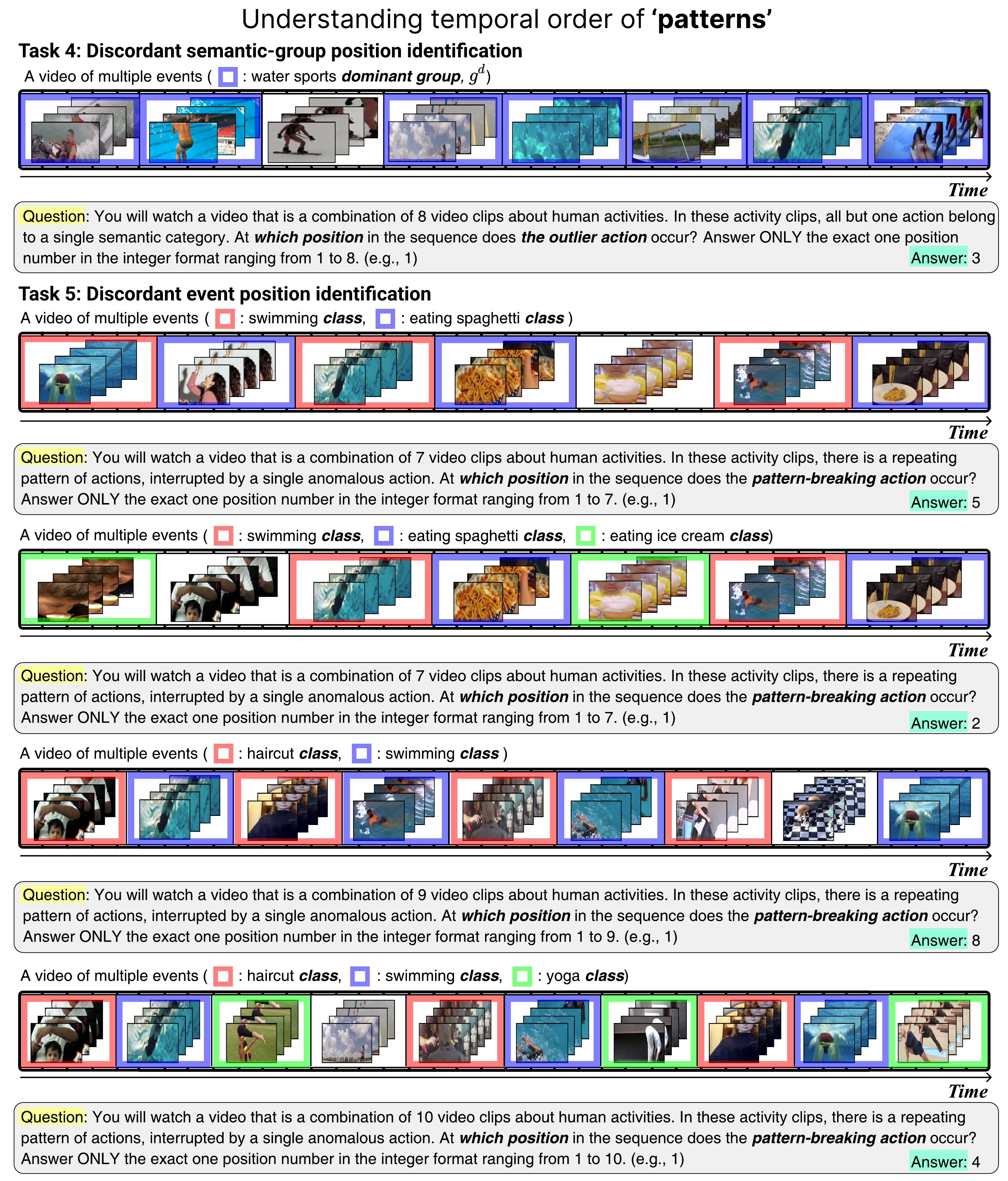}
    \caption{\textbf{Prompt details employed for understanding temporal order of patterns.} We decompose the pattern-level temporal order understanding into two sub-tasks: discordant semantic-group position identification task and discordant event position identification task. We divide discordant event position identification task into four variations ($s_1s_2s_1s_2s_1s_2$ + x, $s_1s_2s_3s_1s_2s_3$ + x, $s_1s_2s_1s_2s_1s_2s_1s_2$ + x, $s_1s_2s_3s_1s_2s_3s_1s_2s_3$ + x) according to the temporal patterns in video composition.
    }
    \vspace{-1em}
    \label{fig:supp_prompt_pattern}
\end{figure*}

Figures~\ref{fig:supp_prompt_event} and \ref{fig:supp_prompt_pattern} illustrate the complete set of task prompts from our \benchmark benchmark. 
Each task is accompanied by answer prompts to guide VLMMs in generating appropriately formatted responses.


\begin{figure}[t]
    \centering
    \includegraphics[width=1\linewidth]{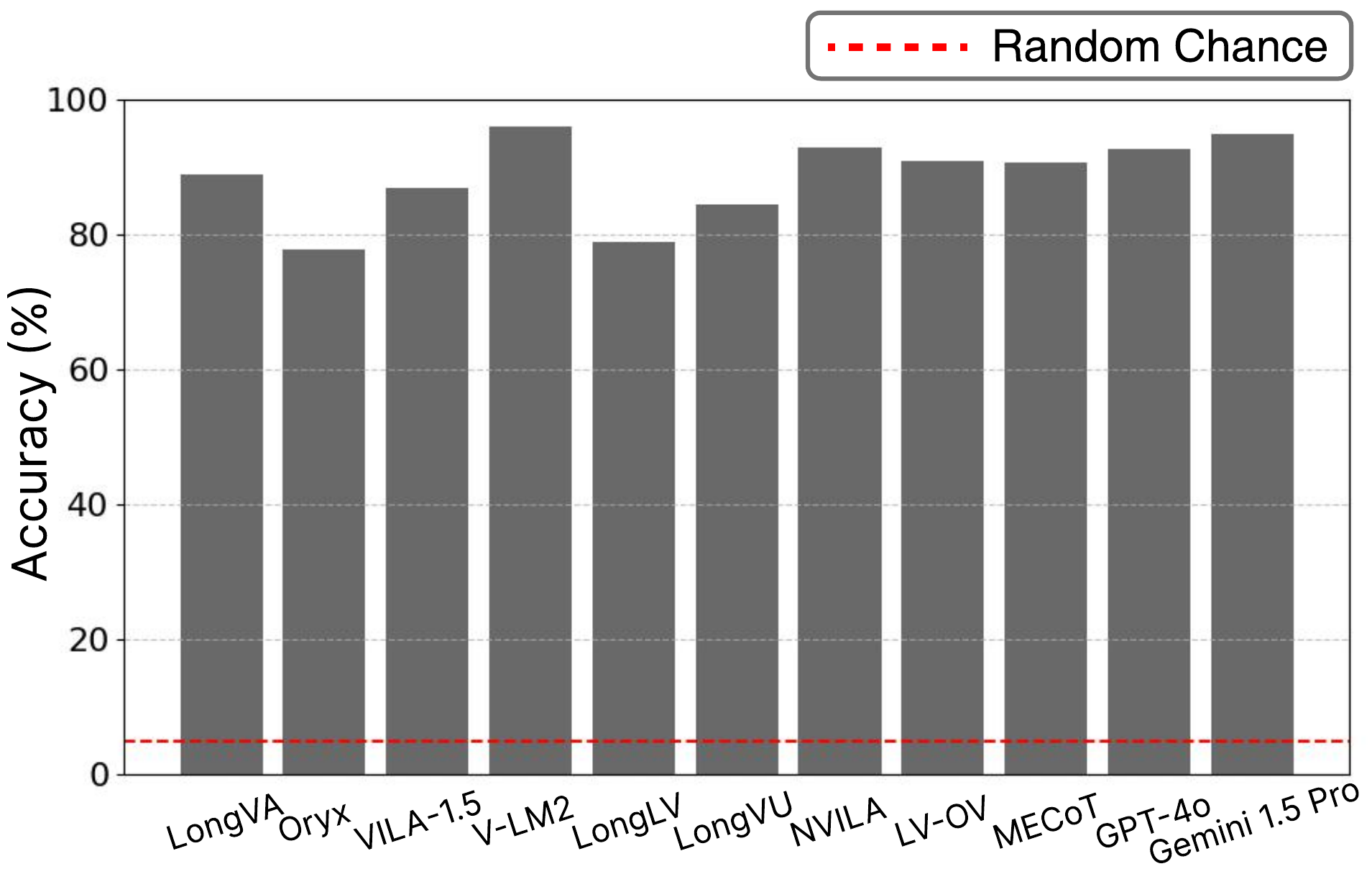}
    \vspace{-0.5em}
    \caption{\textbf{Single-event recognition accuracy on various VLMMs}.
    We conduct experiment on 11 tested VLMMs, including our proposed \method, on single event recognition.
    The VLMM is provided with 20 candidate event categories, and asked to choose the most likely event that is actually happening in the video.
    We report `Accuracy' on 2400 videos sampled from the Kinetics-700 validation set.
    The VLMM could correctly recognize each events separately, achieving 90\% accuracy on average.
    `Random Chance' represents the accuracy of randomly selecting the correct answer.
    }
    \label{fig:supp_single_event}
    \vspace{-1em}
\end{figure}

\section{Single Event Recognition Capability in Various VLMMs}
\label{app:single_event_recognition_capabilities}
Before evaluating VLMMs' ability to understand multiple events in videos, we first assess their single event recognition capabilities. This preliminary experiment ensures that models can accurately recognize individual events.

In this task, each VLMM is tasked with identifying the single major action in the video from a set of 20 candidate action categories $\mathcal{C}^e$. For each VLMM, we provide video clips along with 20 action category labels - the correct class of the video clip plus 19 randomly selected classes - to evaluate their classification accuracy.
We set $|\mathcal C^e| = 20$ to evaluate the single-event recognition capabilities required by our \benchmark benchmark while covering diverse action categories.
For this experiment, we randomly sampled 2,400 videos from the Kinetics-700 validation set.
As shown in Fig.~\ref{fig:supp_single_event}, the VLMMs achieve mean accuracy rates above 90 $\%$, indicating strong proficiency in single-event recognition.

\section{Details about \method}
\label{app:details_method}
\subsection{Constructing Multi-Event Video Description Dataset}
Existing instruction-tuning datasets~[\bmp{32, 37}] often compress multi-event scenarios into single comprehensive summary, losing the ordered information of fine-grained events.
For instance, a video containing multiple distinct actions (entering a room, sitting, and typing) may be simplified as `person working at a desk.', removing key temporal dependencies of actions.
To address this limitation, we construct a multi-event description dataset using a two-stage approach, as illustrated in~Fig.~\bmp{4}-(a).

We first sample 56k short videos from the Kinetics-700 dataset~[\bmp{37}] in training split, then create synthetic video sequences by temporally combining these clips into sets of 3, 4, 5, 6, and 8 segments.
To generate multi-event video descriptions for these synthetic sequences, we follow a two-stage process.
First, a LLaVA-OneVision-7B~[\bmp{15}] produces detailed individual descriptions for each video segment. 
These individual descriptions are then integrated into a coherent narrative with GPT-4o-mini~[\bmp{31}]. This approach results in 120k temporally enriched descriptions for our synthetic videos. These descriptions are then used to fine-tune LLaVA-OneVision-7B, forming the foundation of \method's base model, $\mathcal{M}$, which serves as the backbone for multi-event temporal reasoning.

\subsection{Chain-of-Thought Reasoning}
Although multi-event instruction fine-tuning helps temporal understanding, \emph{explicitly} articulating the reasoning process is essential for recognizing and interpreting events in multi-event videos~[\bmp{45, 53}].
Hence, \method uses a CoT inference strategy~[\bmp{1, 53, 54}], as shown in Fig.~\bmp{4}-(b). 
We obtain our \method by fine-tuning a pre-trained VLMM (7B)~[\bmp{15}] with the above synthetically generated data, and adopting the chain-of-thought in the inference stage. 
Starting from our fine-tuned foundation model, $\mathcal{M}$, CoT explicitly structures the reasoning process by guiding the model through sequential event analysis step by step.

Specifically, given a multi-event video $V$, the fine-tuned $\mathcal{M}$ first generates a chronological \emph{video context} $d$ using a generation prompt $p_{\text{gen}}$:
\begin{equation}
    d = \mathcal{M} ( p_{\text{gen}}, V ).
\end{equation}
Then, the query prompt $p_{\text{query}}$ is concatenated with the generated context to predict $y$:
\begin{equation}
    y = \mathcal{M} \left( p_{\text{query}}||d, V \right).
\end{equation}

This two-step process helps the model capture the temporal structure of multi-event videos more effectively.
As demonstrated in Tab.~\bmp{6}, \method shows significantly improved capabilities in temporal reasoning and event sequence understanding compared to the baseline model.


\end{document}